\documentclass[10pt,twocolumn,letterpaper]{article}

\usepackage[pagenumbers]{cvpr} 

\usepackage{graphicx}
\usepackage{amsmath}
\usepackage{amssymb}
\usepackage{booktabs}

\usepackage{multirow}
\usepackage{algorithm}
\usepackage{algpseudocode}
\usepackage{enumitem}
\usepackage{hhline}
\setlist[itemize]{noitemsep}

\usepackage[pagebackref,breaklinks,colorlinks]{hyperref}

\usepackage[capitalize]{cleveref}
\crefname{section}{Sec.}{Secs.}
\Crefname{section}{Section}{Sections}
\Crefname{table}{Table}{Tables}
\crefname{table}{Tab.}{Tabs.}
\begin{document}

\title{RADU: Ray-Aligned Depth Update Convolutions for ToF Data Denoising}

\author{Michael Schelling\\
Ulm University\\
\and
Pedro Hermosilla\\
Ulm University
\and
Timo Ropinski\\
Ulm University
}
\maketitle

\begin{abstract}
    Time-of-Flight (ToF) cameras are subject to high levels of noise and distortions due to Multi-Path-Interference (MPI). 
    While recent research showed that 2D neural networks are able to outperform previous traditional State-of-the-Art (SOTA) methods on correcting ToF-Data, little research on learning-based approaches has been done to make direct use of the 3D information present in depth images. 
    In this paper, we propose an iterative correcting approach operating in 3D space, that is designed to learn on 2.5D data by enabling 3D point convolutions to correct the points' positions along the view direction. 
    As labeled real world data is scarce for this task, we further train our network with a self-training approach on unlabeled real world data to account for real world statistics. 
    We demonstrate that our method is able to outperform SOTA methods on several datasets, including two real world datasets and a new large-scale synthetic data set introduced in this paper.
\end{abstract}

\section{Introduction}
\label{sec:introduction}

Time of Flight (ToF) cameras are devices that are able to capture depth information of a scene by measuring the time the light emitted by the device needs to travel back once intersecting with an object.
In practice, timing the reception of a light impulse requires precise and costly hardware and, as a result, consumer-level ToF cameras perform indirect measurements.
The most common types are so-called Amplitude-Modulated Continuous-Wave (AMCW) ToF cameras, as they are for example used by the Kinect.
The working principle of these AMCW cameras is to emit a periodically modulated light signal and retrieve the phase shift of the received signal, through which the travel time of the light and, consequently, the distance of the object to the camera is given~\cite{hansard2012tof_principles}.
However, the continuous illumination of the scene inevitably leads to Multi-Path Interference (MPI), as not only the direct reflection is received but also indirectly illumination which significantly impairs the depth estimation.
Furthermore, these ToF cameras suffer from low Signal to Noise Ratios (SNR) on dark surfaces and the mixed pixel effect along sharp object edges~\cite{hansard2012tof_principles}.

\begin{figure}
  \centering
   \includegraphics[width=\linewidth]{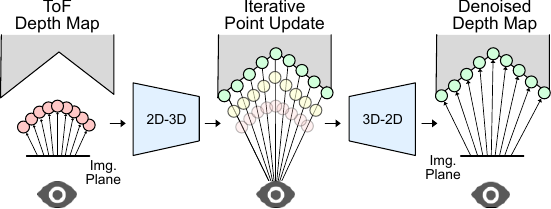}
   \caption{Given an initial ToF depth reconstruction our method projects the problem into a latent 3D space. The 3D point positions are updated iteratively along the camera rays using RADU point convolutions, in order to optimize the final 2D depth prediction.}
   \label{fig:teaser}
\end{figure}

Since deep learning approaches have shown great success in visual computing problems, many works have investigated the capabilities of 2D neural networks to correct ToF depth images~\cite{agresti2018mpiremoval, agresti2019unsupervised, buratto2021deep, su2018end2end, Guo_2018_ECCV, gutierrez2021itof2dtof}.
However, existing methods treat the task of ToF correction as a 2D image problem and do not take into account the explicit 3D information in their computations.
In these works, the depth information is usually used as an input to standard Convolutional Neural Networks (CNN) for images~\cite{marco2017deeptof, agresti2018mpiremoval, agresti2019unsupervised, son2016robotictof, song2018depth}, while the underlying 3D structure is not analyzed.
In this work instead, we propose a new neural network architecture that projects the problem into the 3D domain and makes use of point convolutional neural networks~\cite{hermosilla2018monte} to analyze the erroneous reconstruction and adjust the point positions along the view direction, see \cref{fig:teaser}.
This iterative process makes small changes to the point positions to reduce the error level in between the convolutional layers and improve the final depth reconstruction.
Further, we propose a novel fine-tuning procedure for Unsupervised Domain Adaptation (U-DA) based on self-training methods, to transfer the knowledge acquired by our network from synthetic to real world ToF data. 
The effectiveness of this approach is evaluated on both synthetic and real datasets and proves to be able to outperform existing methods.
Moreover, we introduce a large-scale high-resolution dataset consisting of challenging scenes, containing high MPI levels, materials which produce low SNR captures and objects with high frequency details.
In summary, our contributions are:
\begin{itemize}
    \item A novel architecture for correcting of ToF images in a latent 3D space. 
    \item A two-staged training procedure with a cyclic self-training approach designed to bridge the gap between synthetic and real world ToF images. 
    \item A large-scale high-resolution synthetic ToF dataset containing measurements for scenes with high MPI. 
\end{itemize}
Our synthetic data set, code and trained networks are available at \url{https://github.com/schellmi42/RADU}.

\section{Related Work}\label{sec:rel_work}
In this section we will briefly review existing work related to our approach in different fields.

\noindent \textbf{Learned ToF Correction.}
Beginning with Marco \etal~\cite{marco2017deeptof}, several works on correcting ToF depth images using deep learning have been proposed.
While the former uses a low-frequency ToF depth prediction as input, subsequent work by Agresti \etal~\cite{agresti2018mpiremoval, agresti2019unsupervised, agresti2021unsupervised} greatly improved the reconstruction by considering additional multi-frequency features derived from raw camera measurements to further reduce the error.
The same year, Su \etal~\cite{su2018end2end} proposed a generative approach that generates depths directly from raw measurements.
Instead of predicting a corrected depth directly, Guo \etal~\cite{Guo_2018_ECCV} followed an inverse approach and used a 2D CNN to correct the raw camera measurements prior to the LF2~\cite{lingzhu2016LF2} depth reconstruction algorithm of the Kinect2.
Further, several works on improving on various aspects and applications using machine learning followed, \eg  online calibration using RGB information~\cite{qiu2019deep}, frame rate optimization~\cite{chen2019learning}, power efficiency~\cite{chen2020very}, robotic arm setups~\cite{son2016robotictof} or translucent materials~\cite{song2018depth}, to name a few. 
The aforementioned approaches all use standard 2D CNNs and thus consider the correction problem as an image task.

Recent work aims to estimate the real depth by reconstructing the transient image, \textit{i.e.}~the impulse response of the scene, through learning methods.
Buratto \etal~\cite{buratto2021deep} used the assumption that the direct reflection reaches the camera sensor early and predict the intensity and arrival time of the first two peaks of the impulse response. 
The iToF2dToF method~\cite{gutierrez2021itof2dtof} first predicts ToF depths at various frequencies which are used to estimate the two leading coefficients of the Fourier-Transform of the impulse response.
While transient reconstruction is a promising direction, these methods are, as of now, still very memory demanding, and can thus only query few pixels in parallel, and do not reach the correction capabilities of learned correction approaches. 

In contrast to previous work, we propose to lift the problem into 3D and use 3D point CNNs. 
With this approach we follow the findings of recent research, that neural networks can benefit from latent 3D representations in various tasks, such as object detection~\cite{wang2019pseudo} using 3D voxel CNNs on lidar data, semantic segmentation~\cite{xing20192_5Dconv} using 2.5D convolutions on RGB-D data or even image generation~\cite{nguyen2019hologan}.

\noindent \textbf{Domain Adaptation for ToF Images.}
Real ToF data with labeled ground truth data is only sparsely available, as its collection is rather complex.
Thus various synthetic data sets have been introduced~\cite{agresti2018mpiremoval, marco2017deeptof, Guo_2018_ECCV} to provide the amount of data needed to train deep neural networks.
However the usage of data from a different source, in this case a synthetic simulation, comes at the cost of introducing a domain gap, \textit{i.e.}~real and synthetic images differ in their statistical properties.
Consequently, one major challenge when learning the task of correcting ToF images is bridging the domain gap between synthetic data sets and real world data.

To improve performance on real data Marco \etal~\cite{marco2017deeptof} pre-train an auto-encoder network on unlabeled real world ToF-data, and transfer the encoder layers to a encoder-decoder network for correction.
Other works~\cite{su2018end2end, agresti2019unsupervised} have suggested to use adversarial losses, where a second network acts as an adversarial agent which is trained to distinguish depths generated from real and synthetic data. 
During training the generator is optimized to deceive the discriminator.

Recently, self-training methods, which originated from a student-teacher approach by Hinton \etal~\cite{hinton2015distilling}, have shown great success in various variants of domain-adaptation~\cite{lee2013pseudo, mey2016soft, kumar2020understanding, xie2020innout, zou2019confidence, prabhu2021sentry, zou2018unsupervised, liu2021cycle}.
Building upon these works, we employ a cyclic self-training procedure to adapt our network to real world statistics by generating pseudo-labels for unlabeled real data, after pre-training the network on synthetic data.

\noindent \textbf{Point Cloud Denoising.}
Several learned methods for denoising point clouds have been published~\cite{yu2018ecnet, yifan2019patch, hermosilla2019total, rakotosaona2020pointcleannet, roveri2018pointpronets}. 
In contrast to these approaches, which tackle general noise models, our approach aims to preserve the 2.5D structure of the data by performing 1D corrections. 
Further, MPI errors are not i.i.d., which, to our knowledge, is a typical assumption for unsupervised point cloud denoising methods~\cite{hermosilla2019total}.
\section{Problem Statement}\label{sec:tof_principle}

To estimate the distance $d$ of an object, an AMCW ToF camera emits a light signal $s_e$, which is typically modulated by a sinusoidal periodic function, in the form of
\begin{align}
    s_e(t) &= s_0 \cdot \big(1 + m \cdot \sin(2\pi f t)\big),
\end{align}
where $s_0$ is the average intensity, $m$ is the modulation coefficient, $f$ is the modulation frequency, and $t$ is the time.
For compactness and \textit{w.l.o.g.}~we assume $s_0=1, m=1$.
In the optimal case of only direct reflection the received light $s_r$ is a scaled and phase shifted version of the emitted signal
\begin{align}
    s_r(t) = r\cdot s_e(t - \Delta t) = r \cdot \big(1 + \sin(2\pi f t - \Delta\varphi)\big),\label{eq:signal_received}
\end{align}
where $r$ is the ratio of the light backscattered to the sensor from the surface, and $\Delta\varphi$ is the phase delay after the signal has traveled the distance $2d$, \textit{i.e.}~$\Delta\varphi = 2dc /f$, where $c$ is the speed of light.
The received signal is averaged during the exposure time $\delta t$ at the sensor with a phase shifted version of the emitted signal, resulting in the measurement
\begin{align}
    m_\theta &= \frac{1}{\delta t}\int_{\delta t} s_r(t) \cdot s_e\left(t + \frac{\theta}{2\pi f}\right)\,dt. 
\end{align}
Under the assumption $\delta t \gg 1/f$ the measurement $m_\theta$ can be approximated as~\cite{frank2009theoretical}
\begin{align}
    m_\theta &= I + A \cdot \cos(\Delta\varphi + \theta),
\end{align}
where $I$ is called the intensity and $A$ the amplitude.
By measuring $m_\theta$ for multiple phase offsets $\theta$ the phase shift $\Delta\varphi$ and the distance $d$ can be reconstructed as~\cite{hansard2012tof_principles}
\begin{align}
    \Delta\varphi &= \arctan \left(\frac{\sum_\theta {-\sin(\theta)\cdot m_\theta}}
            {\sum_\theta \cos(\theta)\cdot m_\theta}\right),\label{eq:tof_phase}\\ 
    d &= \frac{c\cdot \Delta\varphi}{4\pi f}.\label{eq:tof_depth}
\end{align}
Due to the periodic nature of the signal, the reconstructed distance $d$ using Eq.~\eqref{eq:tof_depth} is ambiguous for distances larger than $d_{max} = c / (2f)$.
To resolve this so called phase-wrapping the common solution is to acquire measurements for different modulation frequencies $f_k$, which thus also have different maximal distances $d_{max}$.

In practice the received signal $s_r$ is not only the direct reflection after the time $\Delta t$, as in Eq.~\eqref{eq:signal_received}, but a composition of light scattered along various paths $P$ in the scene
\begin{align}
    s_r(t) = \int_{P} r(p) \cdot \big(1 + \sin(2\pi f t - \Delta\varphi(p)\big)\,dp.
\end{align}
While the intensity $r(p)$ can be expected to be low after multiple reflections on an isolated path $p$, the accumulation over all possible paths $P$ leads to the aforementioned notable MPI distortion in the distance recovery, Eq.~\eqref{eq:tof_depth}.
Additionally, ToF-sensors suffer from the common camera noise sources in the form of photon shot noise, thermal shot noise and read noise, which are typically modeled jointly as an additive Gaussian noise on the measurement $m_\theta$.

Although, $d$ from Eq.~\eqref{eq:tof_depth} denotes the distance, it is commonly referred to as ToF depth, as we will do in this work.

\section{Proposed Method}
\label{sec:method}

\begin{figure}
  \centering
   \includegraphics[width=0.9\linewidth]{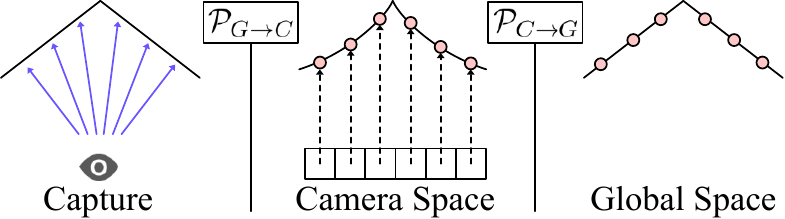}
   \caption{The non-parallelism of camera rays (left) leads to a distortion in the resulting depth image, as the camera measures the distance along the rays (middle).
   By applying the inverse camera projection $\mathcal{P}_{C\rightarrow G}$ the depth map is projected into 3D space, and, in the ideal case, aligns with the scene geometry (right).}
   \label{fig:global_space}
\end{figure}

In this work we propose to incorporate 3D learning techniques to exploit the spatial structure of the scene geometry.
In contrast to 2D convolutions, where spatial relations are encoded implicitly in the pixel grid location, a 3D convolution relies on explicitly given 3D coordinates~\cite{hermosilla2018monte}.
In order to represent the 2.5D data in 3D space we use the inverse of the camera projection $\mathcal{P}_{G\rightarrow C}$ to create an initial spatial position for each pixel, as illustrated in \cref{fig:global_space}.
This allows us to correct the depth in global coordinates, allowing the 3D layers to consider the 3D neighborhoods of the individual pixels, and learn on the actual scene geometry, undistorted by the non-linear camera transformation.

In the case of ToF-data we derive an initial 3D coordinate from the ToF depth given by Eq.~\eqref{eq:tof_depth}. 
Of course this initial depth is inherently erroneous, and we assume that a 3D network would benefit from corrected 3D positions, which lie closer to the scene geometry.
Thus, we propose to iteratively update the 3D positions of the points, enabling the network to optimize the latent point cloud in between the networks 3D layers.
To keep the point position in aligment with the underlying structure given by the pixel position, we restrict the points movement to the respective camera ray.

However, compared to 2D Convolutions, 3D Convolutions are more demanding in compute power and memory consumption.
To reduce this load we embed our proposed 3D layers in a 2D Network and introduce 2.5D pooling layers to reduce the spatial resolution in the 3D convolutions.
This is necessary as for example a $320\times240$ image would result in $76.8k$ points, for comparison common networks for 3D object recognition typically use $1k$ points per model.

In the following we will first define the 2.5D pooling and the RADU convolution layers, before describing our network architecture, the loss function and the self-cycled training procedure for domain adaptation.

\begin{figure*}
    \centering
    \includegraphics[width=\linewidth]{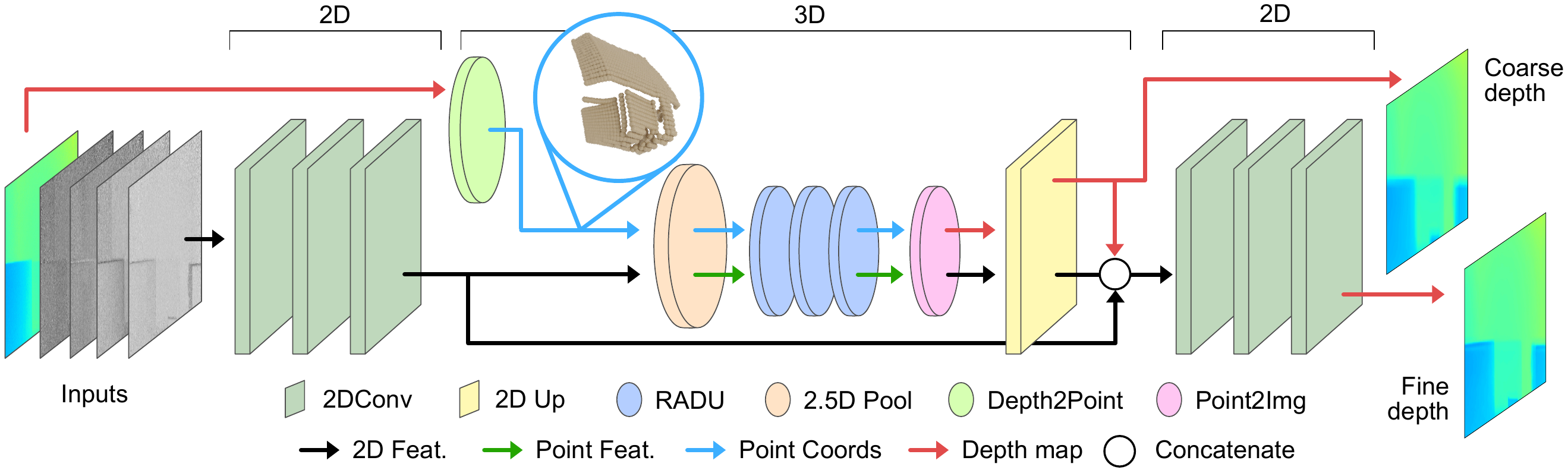}
    \caption{Proposed network architecture. We embed the 3D RADU convolutions in two 2D blocks and use an initial ToF depth given by Eq.~\eqref{eq:tof_depth} to compute initial spatial positions for the 2D features. After the 3D block the point coordinates updated by the RADU layers are projected back to a coarse depth image, which is used as an additional input feature to the following 2D block.}
    \label{fig:net_architecture}
\end{figure*}

\begin{figure}[b]
    \centering
    \includegraphics[width=\linewidth]{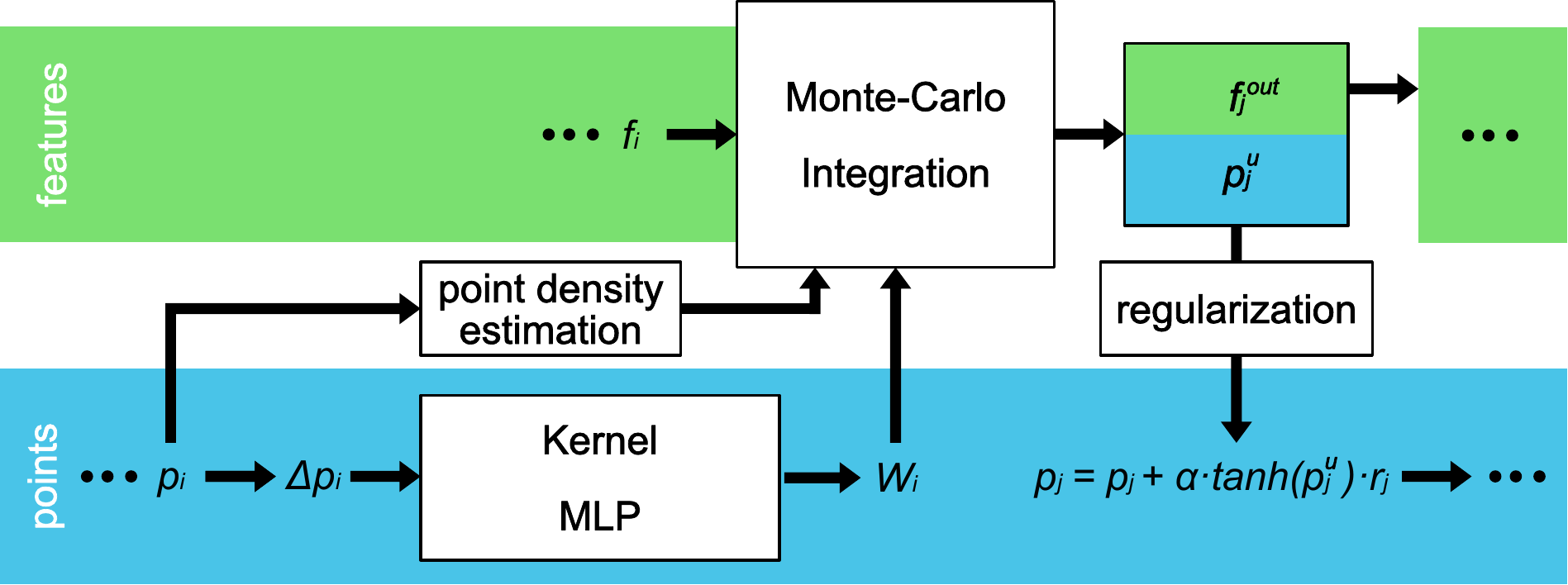}
    \caption{Realization of our RADU convolutional layer based on a 3D Monte-Carlo Convolution. In contrast to standard 3D point convolutions the spatial positions are not only used to predict the kernel weights but are also updated after each convolution.}
    \label{fig:conv}
\end{figure}

\subsection{2.5D Pooling}\label{ssec:2.5D_pool}

Several methods have been proposed for pooling operations on 3D point clouds, \eg Poisson disk sampling~\cite{hermosilla2018monte} and cell average sampling~\cite{thomas2019KPConvsphere}.
However these methods suffer from drawbacks when applied to 2.5D data.
Cell average sampling creates a new point as the mean point in a pre-defined 3D grid, which in general does not lie on the camera ray of a pixel.
Poisson disk sampling chooses a subset of the given points based on their distances in 3D space, making these points still align with the original camera-rays. However this method aims to optimize the distribution of points in 3D space, which can result in non-uniform distributions when projected back into 2D space.

Instead, we propose a 2.5D pooling operation which pools the points in 3D according to their structure induced by the pixel grid, by considering the points' 2D neighborhoods given implicitly through the pixel grid and the camera ray direction associated with the corresponding pixel.
For this we pool only the depth value of the 3D points of a $k\times k$ 2D patch using an order invariant operation, \eg maximum or average pooling, and place a point with the pooled depth on the respective camera ray at the desired reduced resolution $N/k\times N/k$.
The feature of the resulting point is computed by performing a second pooling operation on the features inside the points 2D neighborhood.

This way each resulting point represent a 2D neighborhood of fixed size and aligns with the camera rays.
Note that this is different from (a) pooling the three point coordinates directly, where the result is not necessarily aligned with the camera rays, and (b) 2D pooling on the distance image, as the projections $\mathcal{P}_{G\rightarrow C}$ and $\mathcal{P}_{C\rightarrow G}$ are non-linear, see~\cref{fig:global_space}.
Furthermore, this allows us to exploit existing 2D Pooling operations, dropping the need for costly 3D neighborhood and sampling computations for pooling.

\subsection{Ray-Aligned Depth Update}\label{ssec:RADU}

In order to optimize the point position during the network execution, we propose to extend a given 3D convolution to predict an additional feature channel per layer. 
This value is used to shift the point along the respective camera ray, which we call Ray-Aligned Depth Update (RADU) convolution.
By updating along the camera ray, the point position stays consistent to the 2.5D nature of the point cloud, see~\cref{fig:teaser}.
In order to stabilize the depth updates we regularize the update $p^u$ to the range $(-\alpha, \alpha)$ using a scaled tanh function.
In our experiments we set $\alpha = 0.1$, and provide an ablation study on different values for this hyperparameter in 
the supplementary.
Formally, given a 3D convolution operator $\mathrm{conv}$, a point cloud $\{p_i^{in}\}\subset\mathbb{R}^3$ with associated camera rays $\{r_i\}\subset\mathbb{R}^3$, and input features $\{f_i^{in}\}\subset\mathbb{R}^{C_{in}}$, the RADU convolution on point $p_j^{in}$ with neighborhood $\mathcal{N}_j\subset\mathbb{N}$ is given as
\begin{align}
    (f_j^{out}, p_j^u) &= \mathrm{conv}\left(\{f_i^{in}\}_{i\in\mathcal{N}_j}, \{p_i^{in}\}_{i\in\mathcal{N}_j}\right), \\
    p_j^{out} &= p_j^{in} + \alpha \cdot \tanh(p_j^u) \cdot r_j,
\end{align}
where $f_j^{out}$ is the output feature of the point $p_j^{out}$.

This extension is independent of the type of 3D-convolution used, in our experiments we implement it based on the Monte-Carlo-Convolution~\cite{hermosilla2018monte}, which is illustrated in \cref{fig:conv} and described in more detail in the supplementary.
This type of 3D convolution has been shown to work well with non-uniformly sampled point clouds, as in our case where the density varies with the distance of the points to the camera, which we also investigate in \cref{ssec:ablation_conv_type}.



\subsection{Network Architecture}\label{ssec:net_arch}

Our network architecture is illustrated in \cref{fig:net_architecture}.
It consists of an initial stack of three 2D convolutions with kernel size $3\times3$, which is followed by a 2.5D pooling layer, using average pooling on both depth and features, with a stride of $8\times8$.
This 2D block is followed by a stack of three 3D RADU convolutions with increasing receptive fields of $0.1\,m, 0.2\,m,$ and $0.4\,m$.
After the 3D block we use bilinear upsampling to increase the spatial resolution of the intermediate features and the updated depth values.
The coarse depth prediction of the 3D block is projected back into 2D using the inverse camera transform.
Both the upscaled depth and features are processed by a second block of three 2D convolutions with kernel size $3\times3$.
We further introduce a skip connection between the 2D blocks.
In between convolutions we use leaky ReLU as a non-linearity.

As input to our method we use the same multi-frequency ToF features used in the works of Agresti \etal~\cite{agresti2018mpiremoval, agresti2019unsupervised}.
That is, given measurements at three different frequencies $f_k$, $k=1,2,3$,  we use five input features, $d_1, d_2 - d_1, d_3 - d_1, A_2 / A_1 - 1,$ and $A_3 / A_1 -1$, where $d_k$ is the ToF depth and $A_k$ is the amplitude at frequency $f_k$.
We follow the argumentation of Agresti \etal~\cite{agresti2018mpiremoval}, that the differences $d_1 - d_k, k=2,3$ encode the frequency dependent influence of MPI on the depth recovery, and the relative amplitudes $A_1 / A_k, k=2,3$ provide information about the strength of the MPI for a given pixel.
The depth $d_1$ further serves as initial ToF depth for the projection $\mathcal{P}_{C\rightarrow G}$ into 3D space. 
Further, we augment the input data using rotation, mirroring and noise, a more detailed description of the data augmentation can be found in the supplementary. 

\subsection{Coarse-Fine Loss}\label{ssec:loss}

The 3D block of our network architecture produces an intermediate coarse depth estimate which is fed as an additional input to the subsequent 2D layers. 
To guide the network to predict an adequate representation of the 3D geometry we optimize both the final output $\hat{d}_{out}$ and the coarse 3D representation of the 3D blocks $\hat{d}_{3D}$:
\begin{align}\label{eq:loss}
    \mathcal{L} = \|d_{gt} - \hat{d}_{out}\|_1 + \|d_{gt} - \hat{d}_{3D}\|_1.
\end{align}
As the coarse depth $\hat{d}_{3D}$ is not predicted by the final layer of the 3D network but is constructed iteratively as
\begin{align}
    (\hat{d}_{3D})_j = \mathcal{P}_{G\rightarrow C}\left(d^{init}_j + \sum_{l=1}^3 \alpha \cdot \tanh(p_{j,l}^u) \cdot r_j \right),
\end{align}
this allows each RADU layer to receive gradients directly from the loss function, preventing vanishing gradients, which is comparable to the influence on gradient flow of skip connections. However, as each layer is optimized to produce correctly denoised depths, this also increases the risk of overfitting.
Ablations on the loss function and the choice of $\alpha$ are provided the supplementary.

\subsection{Unsupervised Domain Adaptation}\label{sec:DA}
\begin{algorithm}
    \caption{Adapted Cycled Self-Training Procedure}\label{alg:cst}
    \begin{algorithmic}
    \Require $n_{cycle}\in\mathbb{N}, p\in[0, 1], D_{real}, D_{syn},\text{ network }N$
        \For{$epoch$}
            \If{$epoch\mod n_{cycle}\equiv0$}
                \State $F_{in} \gets D_{real}$ \Comment{Get real world data}
                \State $\hat{d}_{out} = N(F_{in})$ \Comment{predict, unaugmented}
                \State $S_2 \gets \{d_{gt}:\hat{d}_{out}\}$ \Comment{Save pseudo labels}
            \EndIf
            \For{$training\ step$}
                \If{$rand.unif([0,1])<p$}
                    \State $(F_{in}, d_{gt}) \gets D_{real}$\Comment{Real with pseudo label}
                \Else
                    \State $(F_{in}, d_{gt}) \gets D_{syn}$\Comment{Synthetic with label}
                \EndIf
                \State $(F_{in}, d_{gt}) = augment(F_{in}, d_{gt})$
                \State $(\hat{d}_{out}, \hat{d}_{3D}) = N(F_{in})$
                \State $minimize\ \mathcal{L},\text{~Eq.\eqref{eq:loss}, on }(\hat{d}_{out}, \hat{d}_{3D})$
            \EndFor
        \EndFor
    \end{algorithmic}
\end{algorithm}

To improve the performance of our method on real data we investigate a cyclic self-training procedure, derived from existing self-training methods for other tasks~\cite{liu2021cycle, lee2013pseudo, kumar2020understanding}.
We evaluate a network, which is pre-trained on synthetic data, on unlabeled real data and use the predictions as pseudo labels in the following training phase.
During training we choose randomly between synthetic data with labels and real data with pseudo-labels, to prevent the network from overfitting to the pseudo-labels.
We repeat this process multiple times by updating the pseudo-labels every $n_{cycle}$ epochs.
To avoid providing the exact same input during pseudo label generation and training we create pseudo-labels on unaugmented data and use augmented input during training.
The procedure is summarized in \cref{alg:cst}.

We refrain from training a teacher network on labeled real data for pseudo-label generation to keep the assumption of U-DA and to make our approach applicable to settings where no labeled real data is available.

\begin{figure}[b]
    \centering
    \includegraphics[width=0.9\linewidth]{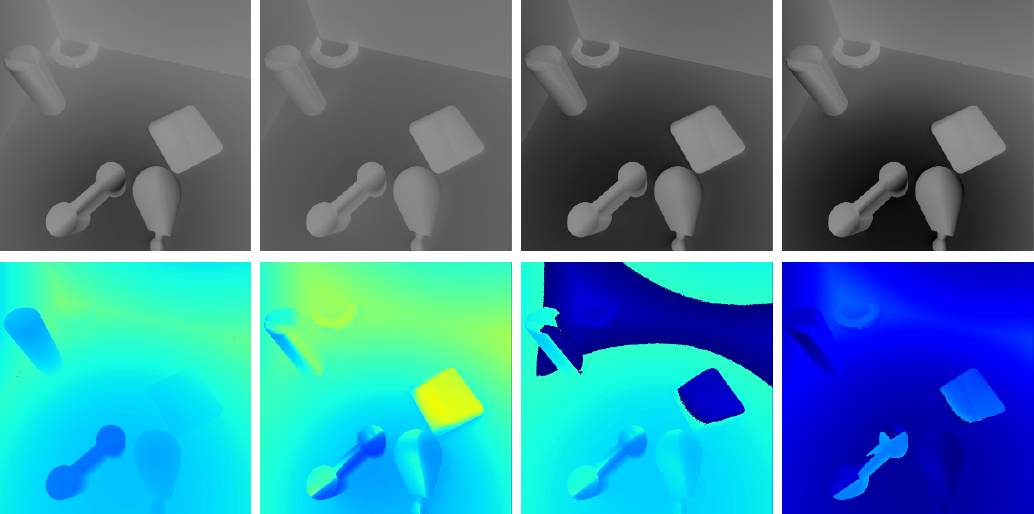}
    \caption{Example scene from our Cornell-Box dataset. The top row shows the four measurements $m_\theta$ at 20MHz, bottom row shows, from left to right, ground truth depth and ToF depths using Eq.~\eqref{eq:tof_depth} at 20MHz, 50MHz, and 70MHz, with phase wrapping.}
    \label{fig:dataset}
\end{figure}

\begin{figure*}
    \centering
    \includegraphics[width=\linewidth]{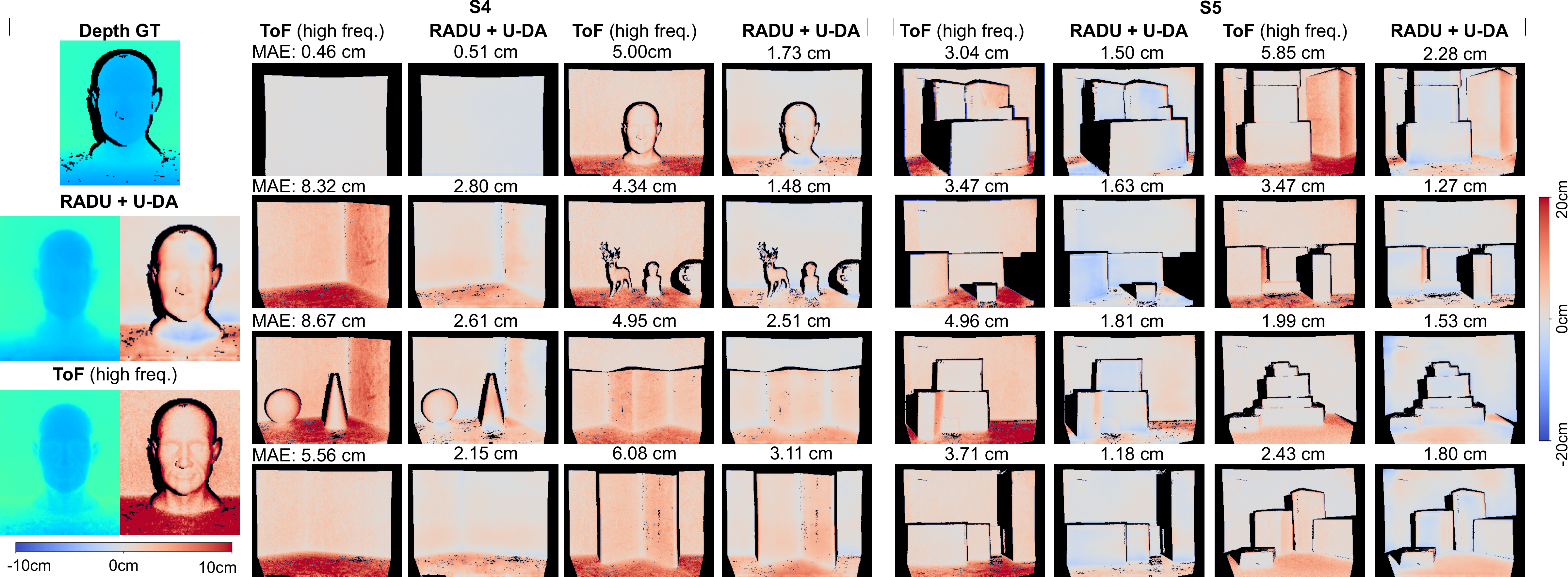}
    \caption{Depth error maps with MAE on the real datasets S4 (left) and S5 (right) of our method using both RADU convolutions and cyclic self-training for unsupervised domain adaptation. 
    Areas were no ground truth depth is available are masked in black. 
    The zoom in on the left displays both depth and error, with enhanced color coding on the error.}
    \label{fig:real_results}
\end{figure*} 

\section{Cornell-Box ToF Dataset}\label{sec:dataset}
To allow a more broad evaluation of our method, we generate a new large scale synthetic dataset.
As recent advances in ToF-camera hardware~\cite{miller2020large} could allow for higher resolution ToF sensors in the future, we render our dataset at a comparably high resolution of $600\times600$ pixels.
Inspired by Miller \etal~\cite{miller2020large} we simulate the properties of a RaspBerry Pi 3 camera equipped with an EAM for modulation.

\begin{table}[b]
  \centering
  \begin{tabular}{lccccl}
    \toprule
    Dataset & Type & GT  & $m_\theta$ & Size &  Resolution\\
    \midrule
    S1~\cite{agresti2018mpiremoval} & Syn. & Yes & No & 54 & 320$\times$240 \\
    S2~\cite{agresti2019unsupervised} & Real & No & No & 96 & 320$\times$239 \\
    S3~\cite{agresti2019unsupervised} & Real & Yes & No & 8 & 320$\times$239 \\
    S4~\cite{agresti2019unsupervised} & Real & Yes & No & 8 & 320$\times$239 \\
    S5~\cite{agresti2018mpiremoval} & Real & Yes & No & 8 & 320$\times$239 \\
    \midrule
    FLAT~\cite{Guo_2018_ECCV} & Syn. & Yes & Yes & 1.2k & 424$\times$512 \\
    \midrule
    Cornell-Box & Syn. & Yes & Yes & 21.3k & 600$\times$600\\
    \bottomrule
  \end{tabular}
  \caption{Properties of the datasets considered in our experiments.}
  \label{tab:datasets}
\end{table}

To create a challenging setup for ToF correction we generate 142 scenes inspired by the Cornell box layout~\cite{goral1984modeling}, to ensure high levels of MPI.
We render each scene from 50 viewpoints with 3 different material properties, including dark materials for low SNR values, which results in 21.3k different renderings.
The data is split into 116 training, 13 validation, and 13 test scenes.
Each rendering is processed to simulate raw measurements for sinusoidal modulations at 20MHz, 50MHz, and 70MHz at phase offsets at $0, \pi/2, \pi,$ and $3\pi/2$, resulting in 12 correlation measurements.
To our knowledge our dataset exceeds existing ToF datasets in both size and resolution, see Table~\ref{tab:datasets}.
We refer the reader to the supplementary for further details about the dataset.

\section{Experiments}\label{sec:experiments}


First we test our method on the established data sets of Agresti \etal~\cite{agresti2018mpiremoval, agresti2019unsupervised}, who provide a synthetic and several real data sets for the SoftKinect camera, containing both MPI and shot noise, to evaluate our method including U-DA.
We refer to these data sets using the same notation as previous authors~\cite{agresti2019unsupervised, buratto2021deep}, see \cref{tab:datasets}.
As the real datasets are rather small, we further evaluate our method on larger synthetic datasets, our dataset described in \cref{sec:dataset}, where we additionally consider phase wrapping in the input features, and the FLAT dataset~\cite{Guo_2018_ECCV}, which simulates a Kinect2 sensor, to also investigate the influence of non-sinusoidal modulations of $s_e(t)$, which violate the assumptions from \cref{sec:tof_principle}.

Further we perform an ablation experiment, comparing RADU layers to other convolution layers.
For additional informations about hyperparameter settings and other aspects of the experiments we refer to the supplementary.

\begin{table}[t]
  \centering
  \begin{tabular}{lcccc}
    \toprule
     & \multicolumn{2}{c}{S4} &\multicolumn{2}{c}{S5}\\
     Method & MAE & Relative & MAE & Relative\\
     & [cm] & Error & [cm] & Error \\
    \midrule
    ToF (low freq.) & 7.28 & - & 5.06 & - \\
    ToF (high freq.) & 5.43 & - & 3.62 & - \\
    SRA~\cite{freedman2014sra}$^\dag$ & 5.11 & 94.1\% & 3.37 & 93.1\% \\
    DeepToF~\cite{marco2017deeptof}$^\dag$ & 5.13 & 70.5\%$^\ast$ & 6.68 & 132\%$^\ast$ \\
    +calibration~\cite{agresti2018mpiremoval}$^\dag$ & 5.46 & 75.0\%$^\ast$ & 3.36 & 66.4\% \\
    TIR~\cite{buratto2021deep}$^\dag$ & 2.60 & 47.9\% & 1.88 & 52.0\% \\
    CFN~\cite{agresti2018mpiremoval}$^\dag$ & 3.19 & 58.7\% & 2.22 & 60.5\% \\
    + U-DA~\cite{agresti2021unsupervised} & 2.31 & 42.5\% & 1.64 & 45.3\% \\
    \hline
    RADU & \textbf{1.83} & \textbf{33.7}\% & 2.59 & 71.5\% \\
    RADU + U-DA & 2.11 & 38.8\% & \textbf{1.63} & \textbf{45.0}\% \\
    \hhline{=====}
    RADU + S-DA & 1.89 & 34.8\% & 1.53 & 42.3\% \\
    
    \bottomrule
  \end{tabular}
  \caption{Results of various methods on the real world datasets S4 and S5. Each row reports the MAE and the relative error with respect to the phase unwrapped high frequency ToF depth. ($\ast$: relative to low frequency,  $\dag$: numbers taken from Buratto \etal~\cite{buratto2021deep})}
  \label{tab:results_ToF_correction}
\end{table}

\subsection{Experiment 1: Real World Data}\label{ssec:real_world_experiment}
The dataset S1-S5 contain intensities, amplitudes and phase unwrapped ToF depths.
Similar to Agresti \etal~\cite{agresti2019unsupervised} we train our network on the synthetic dataset S1 and use the real dataset S3 for validation. 
In the second stage we use the unlabeled real dataset S2 in our cycled self-training procedure for U-DA.
We evaluate our method with and without U-DA on the two real data sets S4, with lower MPI levels but more detailed objects, and S5, a 'box' dataset with higher MPI levels but fewer details.
We compare to the Coarse-Fine-Network (CFN), with~\cite{agresti2021unsupervised} and without U-DA~\cite{agresti2018mpiremoval}, DeepToF~\cite{marco2017deeptof}, Transient Image Reconstruction (TIR)~\cite{buratto2021deep}, and the non-learned Sparse Reconstruction Analysis (SRA)~\cite{freedman2014sra}. The Mean Absolute Error (MAE) and the remaining relative error are reported in  \cref{tab:results_ToF_correction}.

The combination of our RADU network with cycled self-training for U-DA outperforms existing approaches on both datasets and successfully removes noise and MPI from the unwrapped ToF depth images, as can be seen in \cref{fig:real_results}.
To evaluate the stability of the cyclic self-training we repeat the U-DA of our method 10 times and measured a standard deviation of the MAE at 0.072cm on S4 and 0.021cm on S5, which indicates that the proposed U-DA approach is stable.
During our experiments, we noticed that the RADU's performance correlates on the sets \{S1, S4\} and \{S2, S3, S5\}, which is indicated by the drop of performance on the dataset S4 after U-DA.
For comparison, we also report the results of our method after a Supervised Domain Adaptation (S-DA) using the small labeled real dataset S3 for fine-tuning.

\begin{table}
  \centering
  \begin{tabular}{lcccc}
    \toprule
     & \multicolumn{2}{c}{Cornell-Box} & \multicolumn{2}{c}{FLAT} \\
     Method & MAE & Relative & MAE & Relative\\
     & [cm] & Error & [cm] & Error\\
    \midrule
    ToF(low freq.) & 29.0 & -  & 59.34 & -\\
    ToF(high freq.) & 11.14 & - & - & -\\
    DeepToF~\cite{marco2017deeptof} & 10.17 & 35.1\%$^\ast$ & 23.0 & 38.8\%$^\ast$\\
    CFN~\cite{agresti2018mpiremoval,agresti2019unsupervised} & 3.99 & 35.8\%$^\dag$  & 6.29 & 10.6\%$^\ast$\\
    End2End~\cite{su2018end2end} & 5.99 & 53.8\%$^\dag$  & 6.20 & 10.5\%$^\ast$\\
    \hline
    RADU & \textbf{3.64} & \textbf{32.7}\%$^\dag$ & \textbf{3.31} & \textbf{5.58}\%$^\ast$\\
    \bottomrule
  \end{tabular}
  \caption{Results of various methods on unseen data from our synthetic Cornell-Box dataset and the FLAT dataset. Each row reports the depth MAE and the relative error with respect to a phase unwrapped ToF depth. ($\ast$: relative to low frequency, $\dag$:  relative to high frequency)}
  \label{tab:results_on_synthetic}
\end{table}

\subsection{Experiment 2: Cornell-Box Dataset}\label{ssec:our_data_experiment}
We train instances of the previously mentioned CFN, DeepToF and our RADU network on our Cornell-Box dataset described in \cref{sec:dataset}.
Since our dataset contains raw measurements $m_\theta$, we further compare to the End2End network~\cite{su2018end2end}.
For a fairer comparison, we perform a hyperparameter tuning for each method, we refer to the supplementary for details.
As our dataset is purely synthetic we do not use domain adaptation strategies.

We evaluate on the test images where our method achieves the lowest MAE of the mentioned methods as can be seen in \cref{tab:results_on_synthetic}.
The remaining relative error is comparable to the previous experiment, showing that our network, as well as CFN, are able to handle phase wrapping in the input features.
The iterative correction in the latent 3D space in between the three RADU convolutions is shown in \cref{fig:radu_step}.

\begin{figure}[b]
    \centering
    \includegraphics[width=\linewidth]{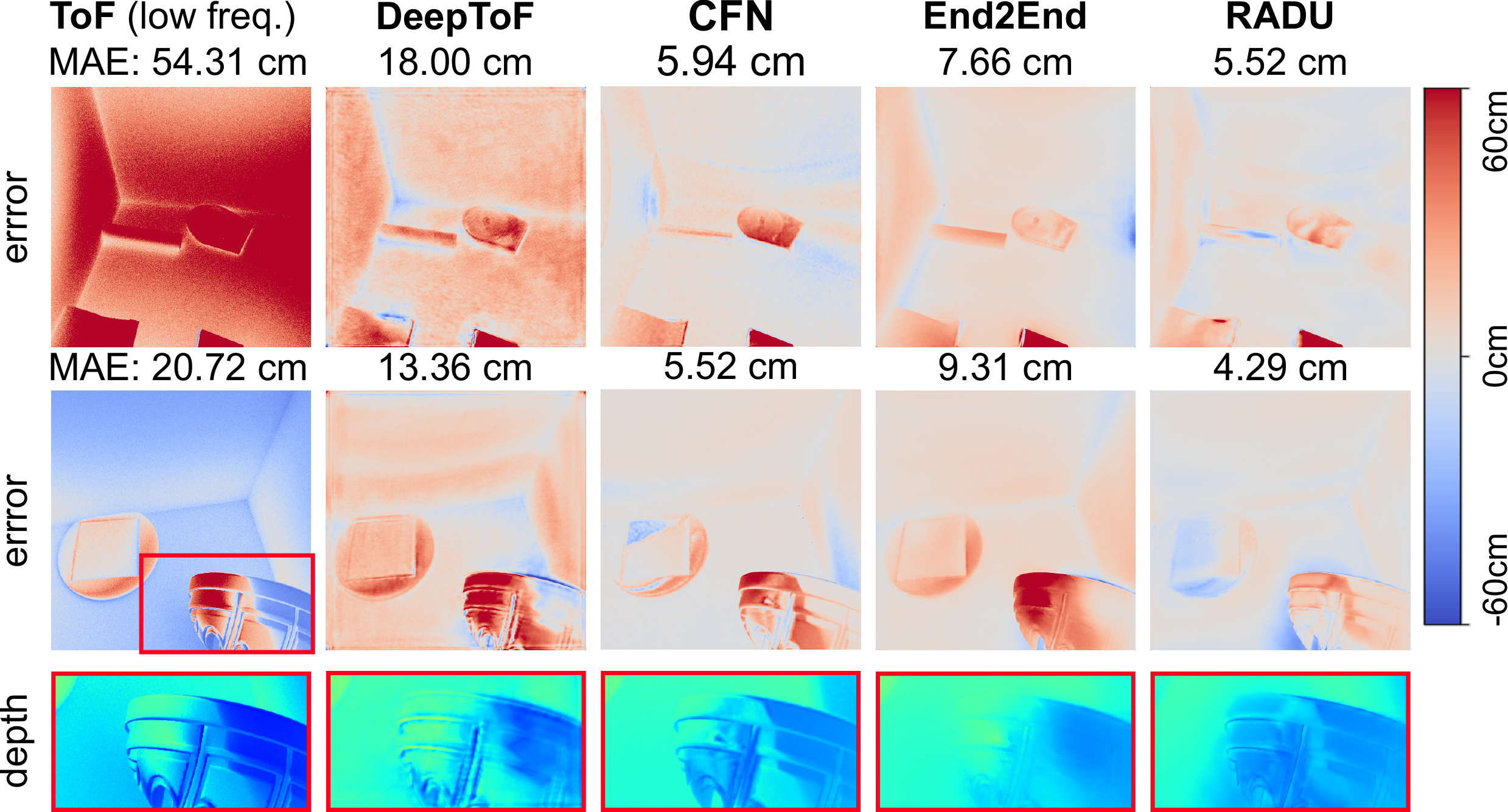}
    \caption{Visual results for two challenging scenes from our Cornell-Box dataset.
    The top scene contains an object with low SNR at the bottom, where all methods fail to retrieve the correct depth. 
    The bottom scene exhibits high MPI, and shows a failure case of our method, where the object boundaries are blurred.}
    \label{fig:ourdata_results}
\end{figure}

\begin{figure}
  \centering
  \includegraphics[width=\linewidth]{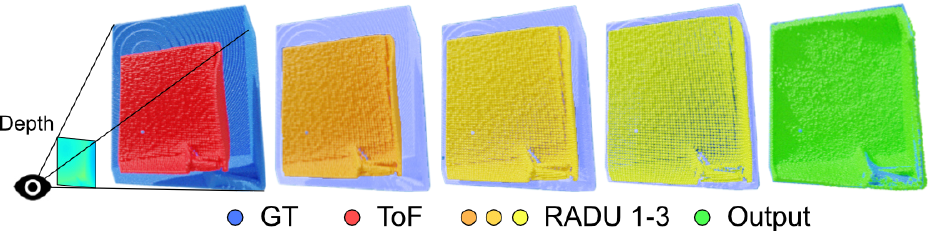}
  \caption{Visualization of the latent point clouds on a corner scene. The depth reconstruction improves after every RADU layer.}
  \label{fig:radu_step}
\end{figure}

However, while our method yields better results than the others, objects with low SNR and scenes with high MPI can still lead to failures in the reconstruction, as shown in \cref{fig:ourdata_results}. 



    

\subsection{Experiment 3: FLAT Dataset}\label{ssec:FLAT_experiment}

In a third experiment we train our method on the FLAT dataset which contains nine raw correlations for three frequencies, simulating the Kinect2. 
Unlike in the previous cases the low frequency signal is non-sinusoidal~\cite{Guo_2018_ECCV} which introduces additional distortions in the depth estimation using Eq.~\eqref{eq:tof_depth}.
A further challenge is the domain gap between the training data, to a large part consisting of images of isolated floating objects and thus low MPI levels, and the test data, containing complete scenes.
We compare our method to the same methods as in the previous experiment, again performing a hyperparameter search for each method.

\begin{figure}
    \centering
    \includegraphics[width=\linewidth]{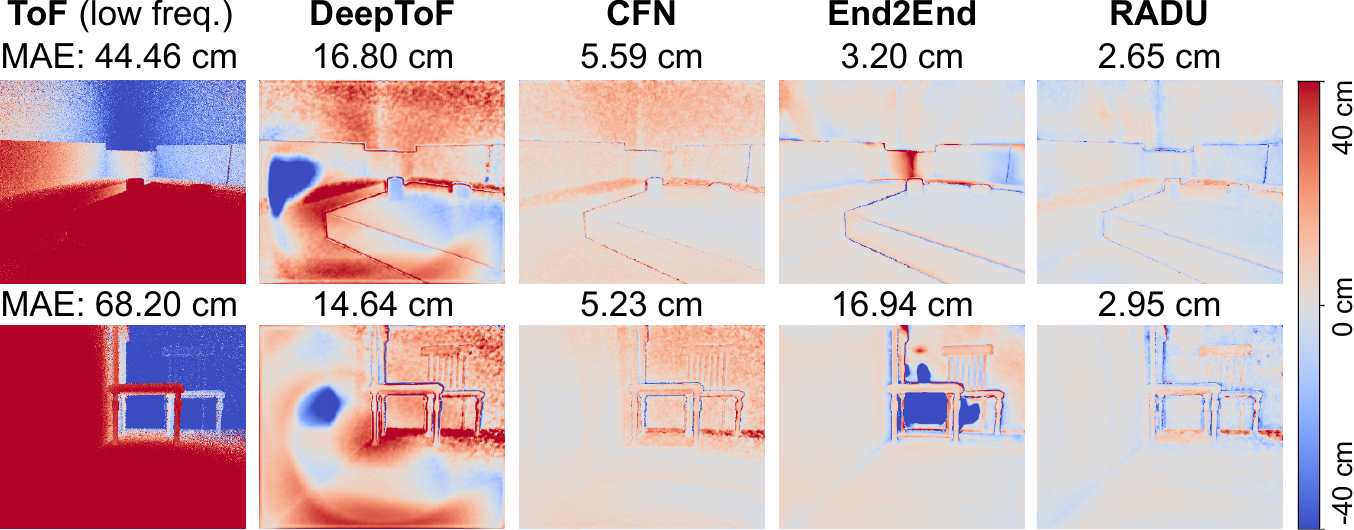}
    \caption{Error maps on the FLAT dataset. 
    The top row shows an example where both RADU and End2End produce an almost error free prediction.
    The bottom row shows a failure case of End2End.}
    \label{fig:flat_results}
\end{figure}

We evaluate on the 120 test images, where our method achieves the lowest MAE compared to the other learned methods, see \cref{tab:results_on_synthetic}.
In \cref{fig:flat_results} we show error images for the different methods.
We recognize that End2End can produce competitive results in some cases but has a tendency to create artifacts, which we assume to stem from the domain gap between training and test data.

\begin{table}[b]
    \centering
    \begin{tabular}{lcc}
        \toprule
         Method & Training (S1) & Validation (S3) \\
         & MAE [cm] & MAE [cm] \\
         \midrule
         2D Conv  & 9.68 & 2.72\\
         2.5D Conv~\cite{xing20192_5Dconv} & 8.79 & 2.61 \\
         3D KPConv~\cite{thomas2019KPConvsphere} & 10.86 & 4.00\\
         3D PointConv~\cite{wu2019pointconv} & 10.49 & 3.42\\
         3D MCConv~\cite{hermosilla2018monte} & 8.38 & 2.51\\
         3D MCConv + RADU & 7.87 & 2.28 \\
         \bottomrule
    \end{tabular}
    \caption{Results of our network architecture with different layer types in the latent block. We report MAE on training (synthetic) and validation (real) data after hyperparameter optimization.}
    \label{tab:ablation_convolution_type}
\end{table}
\subsection{Ablation: Latent 3D Representation}\label{ssec:ablation_conv_type}

To validate the benefits of our 3D RADU convolutions we evaluate the performance when replacing the RADU convolutions with 2D, 2.5D or 3D point convolutions.
In detail we compare to 2.5D convolutions~\cite{xing20192_5Dconv}, which performs three 2D convolutions on separate depth ranges, KPConv~\cite{thomas2019KPConvsphere} which uses spherical kernel points to represent the 3D kernel function, MCConv~\cite{hermosilla2018monte}, which uses kernel MLPs and a density estimation, and PointConv~\cite{wu2019pointconv} which uses a kernel MLP with a learned density estimation.
We conduct a smaller experiment without domain adaption, using S1 for training and S3 for validation.
For a fair comparison we conduct a hyperparameter search for each method, and report results of the validation MAE in \cref{tab:ablation_convolution_type}.

The results show that latent 3D information can help to improve the performance, but the choice of the convolution type is critical.
Both the 2.5D convolution and MCConv yield better results than the 2D convolution, on both training and validation data, outperformed only by our proposed RADU extension of the MCConv layer.

\section{Limitations}\label{sec:limitations}
While we covered several error sources present in ToF data in our experiments, we did not investigate motion artifacts that occur in dynamic scenes~\cite{hansard2012tof_principles}, which would require additional considerations in the network architecture as shown in previous works~\cite{Guo_2018_ECCV, qiu2019deep}.
However, we believe that the latent corrected 3D point cloud representation of our network can potentially be used to improve image alignment.
Further, the results on our Cornell-Box dataset indicate that high MPI, low SNR and fine details can have a drastic impact in the depth estimation, not only for our method but for all evaluated methods.
Finally learned methods, including ours, are trained for specific sensor data and are thus not able to generalize to for example different modulation frequencies. We believe learned methods which are able to treat sensor properties as input parameters, as in traditional approaches~\cite{freedman2014sra}, is a promising line of research. 

\section{Conclusion}\label{sec:conclusion}
In this work we presented an extension of unstructured 3D convolutions for ToF correction, which exploits structured information from depth images to iteratively correct point clouds in 3D space. 
The experiments indicate that latent 3D representations improve the correction capabilities of neural networks for various error sources present in ToF data. 
Further, we demonstrated that cyclic self-training using pseudo-labels can effectively be used for unsupervised domain adaptation on ToF data and, applied to our network, outperforms existing methods on two real data sets. 

While we demonstrated our RADU layers in the context of ToF correction, they can in principle be applied to any task where 2.5D information is present, in order to benefit from both a latent 3D representation and an iterative correction of the former.

\section{Acknowledgements}
This project is financed by the Baden-Württemberg Stiftung gGmbH.
We thank Markus Miller and Rainer Michalzik for their insights on the AMCW ToF working principle and the noise simulation.



{\small
\bibliographystyle{ieee_fullname}
\bibliography{egbib}
}
\ \\\newpage
\appendix
\section{Further Insights on ToF Working Principle}

A convenient formulation for the sinusoidal part $v$ of the measurements $m_\theta$ , is to represent it in the complex plane $\mathbb{C}$ via
\begin{align}
    v &= A\cdot e^{i\Delta\varphi}, \\
    Re(v) &= \sum_\theta -\sin(\theta)\cdot m_\theta, \\
    Im(v) &= \sum_\theta cos(\theta)\cdot m_\theta.
\end{align}
Which allows to express the measurements $m_\theta$ as \cite{frank2009theoretical}
\begin{align}
    m_\theta &= I + A \cos(\Delta\varphi + \theta), \label{eq:tof_measurement}\\
    &= I + Re\left(v\cdot e^{i\theta}\right),
\end{align}
Using this notation the amplitude $A$, the intensity $I$ and the the phase delay $\Delta\varphi$ can be recovered as
\begin{align}
    A &= \|v\|_2.\label{eq:amplitude}\\
    I &= \frac{1}{N}\sum_\theta m_\theta, \label{eq:intensity}\\
    \Delta\varphi &= arg(v)\\ 
    &= \arctan \left(\frac{Re(v)}{Im(v)}\right),\label{eq:tof_phase_supp}
\end{align}
where $N$ is the number of phase shifted measurements $m_\theta$.
Note that this is not an exact solution but a least-squares optimal solution \cite{frank2009theoretical}.

Given measurements at two different frequencies $f_1, f_2$ a computationally cheap solution \cite{hansard2012tof_principles} to unwrap the distances is by minimizing
\begin{align}
    \underset{m, n}{\min} \big| d(f_1) + m \cdot d_{max}(f_1) - d(f_2) + n\cdot d_{max}(f_2) \big|. \label{eq_tof_unwrap}
\end{align}
We use this approach to compute the phase unwrapped high frequency ToF depth in the experiment on our Cornell-Box dataset.

\section{Extension of MC-Convolutions with RADU}

\noindent\textbf{Monte-Carlo Convolution} The Monte-Carlo point convolution of Hermosilla \etal~\cite{hermosilla2018monte} approximates the convolution of two continuous functions, the features $f$ and the kernel function $g$, where $f$ is only known at the discrete positions $\{p_i\}$, the point cloud, using the Monte-Carlo numerical integration.
Formally, let a point cloud $\{p_i^{in}\}\subset\mathbb{R}^3$, input features $\{f_i^{in}\}\subset\mathbb{R}^{C_{in}}$ and a convolution radius $r\in\mathbb{R}$ be given. Further let $\mathcal{N}_j$ denote the neighborhood of $p_j^{in}$, given as 
\begin{align}
    \mathcal{N}_j = \left\{i\ :\ \|p_j-p_i\|_2 \leq r\right\}\subset\mathbb{N}.
\end{align}
Then the Monte-Carlo convolution on point $p_j$ with radius $r$ is defined as
\begin{align}
    (f\ast g)  (p_j) &:= |\mathcal{N}_j|^{-1} 
        \sum_{i\in\mathcal{N}_j} \frac{f_i \cdot g\left(\frac{p_i - p_j}{r}\right)}{pde(p_i\ |\ p_j)},\label{eq:mcconv}
\end{align}
where $pde(p_i\ |\ p_j)$ denotes a point-density estimation of $p_i$ inside the receptive field of $p_j$, and the kernel function $g:\mathbb{R}^3 \rightarrow \mathbb{R}^{C_{in}\times C_{out}}$ is represented implicitly using one or multiple MLPs.
In analogy to 2D convolutions, the evaluation of $g$ on the relative position $(p_i - p_j)/r$ yields the convolution weight matrix $W_{ij}$.

\noindent\textbf{RADU} To predict the additional point update $p_j^{u}\in\mathbb{R}$ we extend $g$ to predict an additional output channel, \textit{i.e.}~
\begin{align}
    g:\mathbb{R}^3 &\rightarrow \mathbb{R}^{C_{in}\times C_{out}} \times \mathbb{R}^{C_{in}\times 1} \notag\\
    &\quad \simeq \mathbb{R}^{C_{in}\times (C_{out} + 1)}, \\
    \frac{p_i - p_j}{r}&\mapsto\left(W_{ij}, W^u_{ij}\right).
\end{align}
Using this kernel function in Eq.~\eqref{eq:mcconv}, the dimensionality of the output of the convolution is increased by one, \textit{i.e.}~$(f\ast g)  (p_j) \in \mathbb{R}^{C_{out} + 1}$.
We split the output into features $f_j^{out}\in\mathbb{R}^{C_{out}}$, and the point update $p_j^u$.
To summarize the point update is computed as 
\begin{align}
    p_j^u &:= |\mathcal{N}_j|^{-1} 
        \sum_{i\in\mathcal{N}_j} \frac{f_i \cdot W^u_{ij}}{pde(p_i\ |\ p_j)}.
\end{align}
The final update along the associated camera ray is performed as described in the main paper
\begin{align}
    p_j^{out} &= p_j^{in} + \alpha \cdot \tanh(p_j^u) \cdot r_j.
\end{align}

\noindent\textbf{Computational Complexity of RADU}
Let $C_{in},C_{out}$ be the number of input and output features, respectively. 
The number of parameters $\#P$ of the above described a 3D RADU convolution with hidden dimension $h$ in the kernel MLP is given as 
\begin{align}
    \#P_{3D} &= h\cdot C_{in}\cdot (C_{out}+1) + 3h + C_{out},\\
    & \in\mathcal{O}(h\cdot C_{in}C_{out}).
\end{align}
The \#P of a 2D convolution with kernel size $k$ is 
\begin{align}
    \#P_{2D} &= k^2\cdot C_{in} \cdot C_{out} + C_{out},\\
    &\in\mathcal{O}(k^2\cdot C_{in}C_{out}).
\end{align}
The leading terms are $h\cdot C_{in}C_{out}$ and $k^2\cdot C_{in}C_{out}$. 
In our RADU network we use a hidden MLP dimension of $h=16$ which places its number of parameters between a 2D convolution with $k=3$ and $k=5$.

\section{Network Architecture - Hyperparameters}
We provide additional parameters of our network architecture for a full description.
Our networks consists of three initial 2D convolutions, followed by a 2.5D pooling layer, three 3D RADU convolutions, a 3D-2D projection with up-scaling and a final stack of three 2D convolutions.

The first three 2D convolutions have feature channels of sizes [64, 64, 128].
The 2.5D pooling layer uses a stride of $8\times 8$ and applies average pooling on both the point depths and the features.
The 3D RADU convolutions use single MLPs with 16 hidden neurons as kernel functions, and a regularization of the point updates with $\alpha=0.1$ m.
The receptive fields are [0.1 m, 0.2 m, 0.4 m] and the feature channels are of sizes [128, 256, 128].
The up-scaling layer uses bilinear interpolation.
The extracted depth of the points is projected back into a 2D depth image and is used as additional input to the following 2D layers.
The final 2D convolutions have feature channels of sizes [64, 64, 1].

In between all convolutional layers, 2D and 3D, we use leaky ReLU with $\alpha=0.1$ as non-linear activation.
All 2D convolutions further have a kernel size of $3\times 3$ and use \emph{'same'} padding.

Finally we use a skip connection between the two 2D blocks with feature concatenation.

\section{Cornell-Box - Dataset Generation}

We generate our dataset using the transient renderer of Jarabo \etal~\cite{Jarabo14transient}, which has been deployed for ToF data generation in previous works \cite{marco2017deeptof, Guo_2018_ECCV}.

\noindent\textbf{Camera Properties}
Inspired by the work of Miller \etal \cite{miller2020large}, which propose an approach to modify standard cameras for ToF captures, we simulate the properties of a RaspBerry Pi 3 camera equipped with an Electro-Absorption Modulator (EAM), which modulates the received signal $s_r$ in front of the camera sensor.

\begin{figure}
    \centering
    \includegraphics[width=\linewidth]{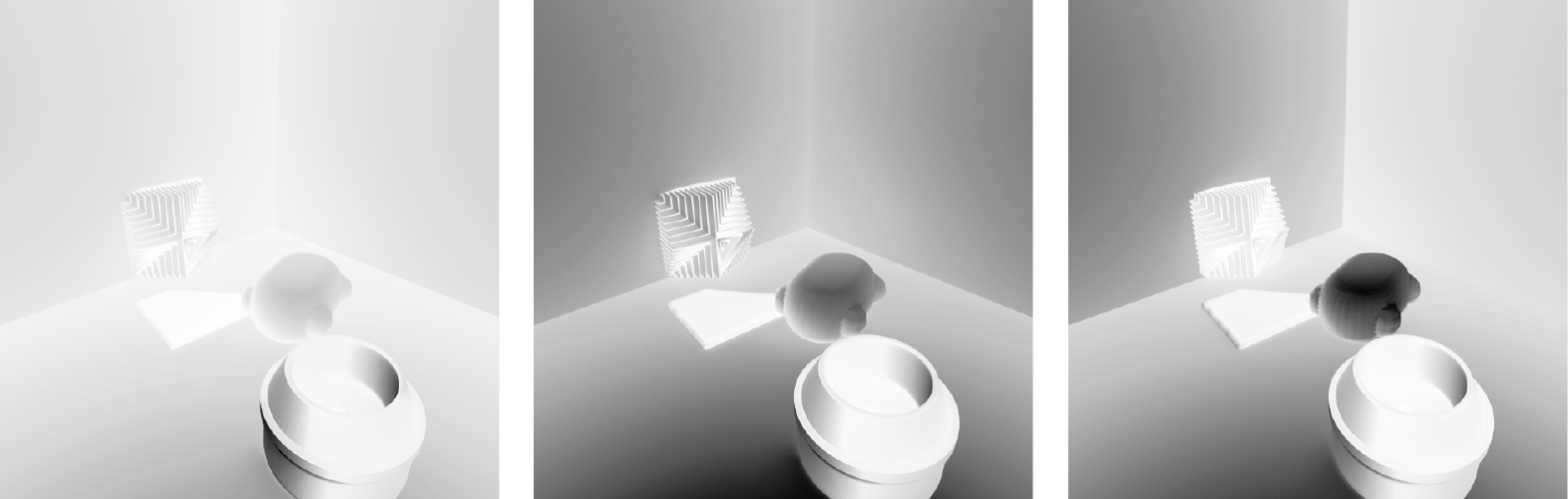}
    \caption{An example scene with three different material properties. The images show the intensity $I$ at a frequency of 20 MHz as given by Eq.~\eqref{eq:intensity}. In the left scene the box material is highly reflective, the right scene contains an object with a dark material.}
    \label{fig:materials}
\end{figure}

\noindent\textbf{Scene Generation}
To ensure challenging scenes with high MPI levels we design our scenes inspired by the Cornell-Box~\cite{goral1984modeling} layout, and place a random number of objects, between 1 and 10, in the scene.
The used objects meshes are taken from a subset of the Thingi10k dataset~\cite{zhou2016thingi10k}, containing 3D models under CC license.
For each 3D object a material property is randomly sampled, the assignment of dark materials allows to simulate regions with lower SNR values.
Further the material of the surrounding box allows, to some degree, control over the level of MPI in the scene.
An example of a scene with different materials is shown in~\cref{fig:materials}.

\begin{figure}[t]
    \centering
    \includegraphics[width=\linewidth]{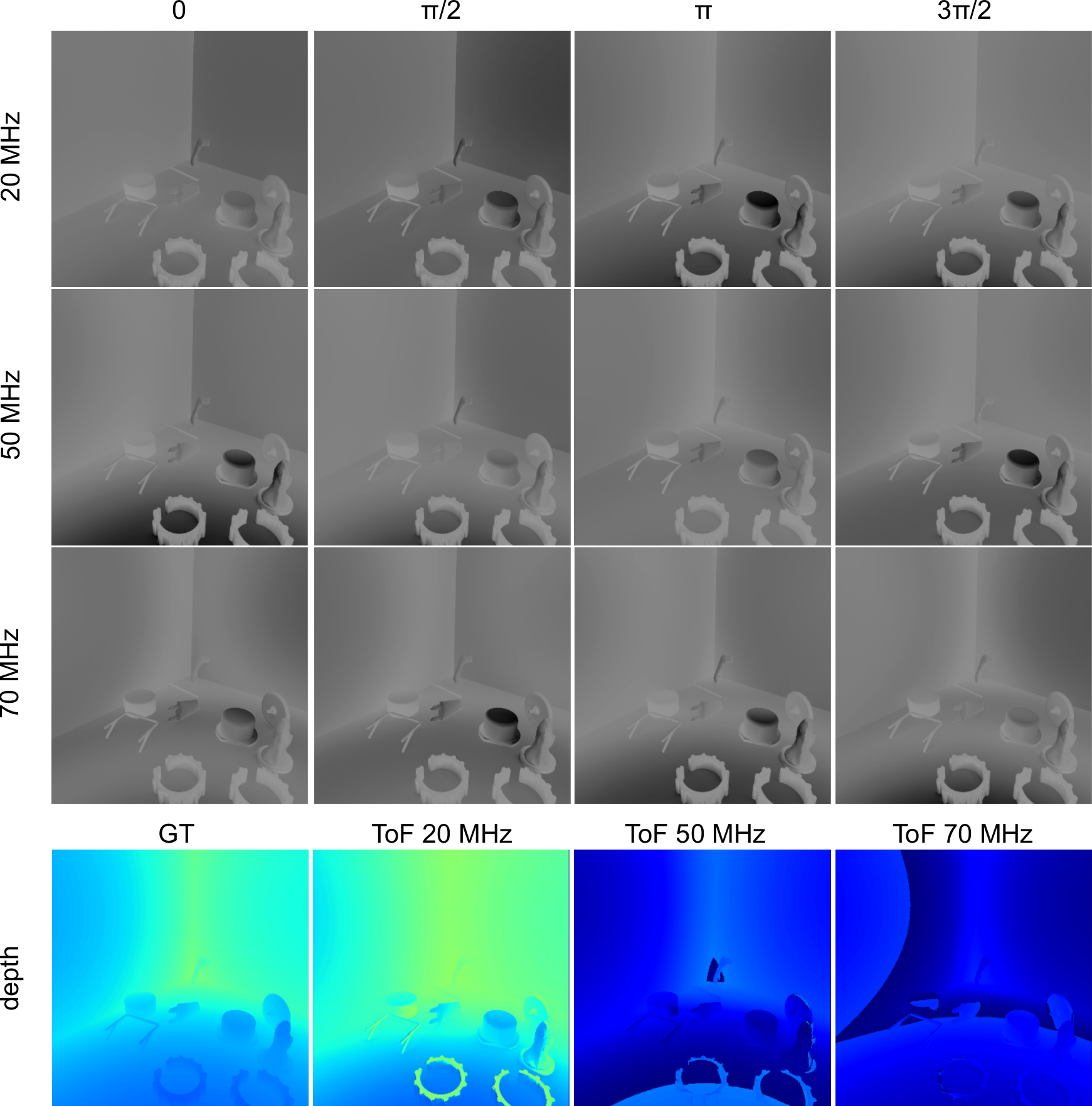}
    \caption{Example scene from our Cornell-Box dataset. Simulated measurements $m_\theta$ for different values of $\theta\in\{0, \pi/2, \pi, 3\pi/2\}$ and frequencies $f\in\{20, 50, 70\}$ are shown in the top rows.
    The reconstructed ToF depths, including phase wrapping, are shown in the bottom row.}
    \label{fig:dataset_example}
\end{figure}

\noindent\textbf{ToF Simulation}
Using the transient renderer we compute the impulse response $h(t)$ of the scene, which contains the received signal $s_r$ in a time resolved format after illuminating the scene with a light impulse.
Given the impulse response the measurements $m_\theta$ can be simulated for different frequencies $f$ and phase offsets $\theta$ using
\begin{align}
    s_r(t) &= h(t) \ast s_e(t),
\end{align}
in the formula from the main paper
\begin{align}
    m_\theta &= \frac{1}{\delta t}\int_{\delta t} s_r(t) \cdot s_e\left(t + \frac{\theta}{2\pi f}\right)\,dt. 
\end{align}
The resulting measurements for an example scene are shown in~\cref{fig:dataset_example}.
Further a scene table can be found at the end of this document in~\cref{fig:scene_table1}, \ref{fig:scene_table2} and \ref{fig:scene_table3}.

\noindent\textbf{Additional Noise}
We simulate the combination of shot noise, thermal noise and read noise as an Additive White Gaussian Noise (AWGN) using the specifications of the RaspBerry Pi 3 camera of Pagnutti \etal~\cite{pagnutti2017RaspPi}, who use the linear noise model of the EMVA Standard 1288~\cite{european2010standard}.
We use the ISO 100 measurements of the former as reference, who measured a gain $K$ of $0.33$ and a Y-intercept $b$ of $-18.4$ on the mean-variance curves
\begin{align}
    \sigma^2 = K\cdot m + b, \label{eq:noise_char}
\end{align}
where $m$ is the mean value across multiple measurements and $\sigma^2$ is the variance of the measurements. 
From Eq.~\eqref{eq:noise_char} we infer the pixel-wise variance $\sigma_{x,y}^2$ dependent on the measurement $m_\theta$ on pixel $(x,y)$ for the AWGN $\mathcal{N}(0, \sigma_{x,y}^2)$.

In our dataset, we provide measurements $m_\theta$ without the AWGN to allow the online generation of AWGN during training for data augmentation, and to allow future researchers to simulate other noise models.

\section{Experiments}
We briefly describe additional details about the experiments, including data augmentation and hyperparameter settings.

The ADAM optimizer\cite{kingma2014adam} is used during backpropagation in all trainings described below.

\subsection{Data Augmentation}

To increase the variety in the datasets we use the following data augmentation strategies on the input features.\\
\emph{Mirroring} Random mirroring along image axes.\\
\emph{Image Rotation} Random rotation by $0^\circ, 90^\circ, 180^\circ, 270^\circ$.\\
\emph{Small Rotation} Additional rotation with a random angle in the range $[-5^\circ, 5^\circ]$. Values outside the image boundaries are interpolated with a \emph{nearest} strategy, as implemented in the preprocessing pipeline of \verb|tensorflow.keras|.\\
\emph{Noise} Additive Gaussian noise with a relative standard deviation of $0.02$.\\
\emph{Random Cropping} In the case of training on image patches we crop random regions of the images every epoch.

We further experimented with the MPI augmentation of Agresti \etal~\cite{agresti2019unsupervised}, but found it did not improve the performance in our experiments.

\subsection{Soft-Kinect - Datasets S1-S5}

Since no raw measurements are available in the dataset, we do not compare to the End2End method in this setting.
While in theory a reconstruction of $m_\theta$ could be done using Eq.~\eqref{eq:tof_measurement}, the error accumulation results in data unfit for training. 

The input resolution of the Soft-Kinect is $320\times240$, which results in a latent point cloud of $1.2k$ points after the 2.5D pooling in our proposed network architecture.
On the real datasets of size $320\times239$ we pad the input to the using reflective padding to the full resolution $320\times240$.

We use the same input features as CFN, by setting $f_1=70\text{MHz}, f_2=20\text{MHz}, f_3=40\text{MHz}$. 
Note, the provided  ToF depth at 70MHz is phase unwrapped in this experiment.

\noindent\textbf{Pre-Training}
Our network is pre-trained on the synthetic dataset S1 using a learning rate of 1e-3 and an exponential learning rate decay of 0.1 every 100 epochs and a batch size of 8. The network converged after 300 epochs.

\noindent\textbf{Cyclic Self-Training}
After pre-training our network on synthetic data, we train the network on the unlabeled real dataset S2 using pseudo-labels as described in the main paper.
In each training step, we choose real examples with a probability of $p=0.5$ and update the pseudo-labels every $n_{cycle}=20$ epochs. We train with a small learning rate of 5e-5 and a batch size of 4 for 100 epochs.

\noindent\textbf{Supervised Domain Adaptation} For comparison we fine-tune our network, after pre-training on synthetic data, using the labeled real dataset S3. We train with a small learning rate of 1e-5 and a batch size of 4 for 100 epochs.

\subsection{RaspBerry-Pi 3 - Cornell-Box Dataset}

We use a resolution of $512\times512$ during training which results in 4096 points in the latent point clouds.
The input features for our network are computed by using $f_1 = 20\text{MHz}, f_2=50\text{MHz}, f_3=70\text{MHz}$.
The network is trained with an initial learning rate of 1e-3 and an exponential learning rate decay of 0.1 every 100 epochs, and a batch size of 4. The network converged after 300 epochs.

\noindent\textbf{DeepToF} includes an Auto-Encoder (AE) training stage on real data for domain adaptation.
As there is no real data in this experiment, we train models with pre-training the AE, as described in the original paper, and training the entire network combined, without pre-training.
We also tune the learning rate with initial learning rates from \{1e-3, 1e-4\} and decay steps from \{50, 75, 100\}, The learning rate in the AE stage is set constant at 1e-4 for 15 epochs, as suggested in the original paper~\cite{marco2017deeptof}.
The learning rate decay is not fully specified in the original paper, we assume a decay of 0.1 every 75 epochs, which matches the authors description.
We use an L2-Loss, a batch size of 16, and the low frequency 20MHz ToF-depth as input as in the original paper.

\noindent\textbf{CFN} was investigated in two papers, we use the more recent version~\cite{agresti2019unsupervised} as reference for our experiments, which predicts the depth directly and does not use additional filtering algorithms.
As no real data is available we drop the unsupervised adversarial part of the training, as with our network architecture. 
The original paper~\cite{agresti2019unsupervised} used a fixed learning rate of 5e-6.
We investigate the static learning rates \{1e-4, 1e-5, 5e-6\} and also in combination with a learning rate decay after \{100, 150\} epochs.
We use a coarse-fine L1-loss and a batch size of 4 as in the original paper~\cite{agresti2019unsupervised}.
The input features for CFN are the same as for our network.

\noindent\textbf{End2End} predicts depths directly from raw correlations, using a generative approach, which we also incorporate into our trainings.
The original network uses a static learning rate of 5e-4 for 50 epochs before decaying the learning rate linearly to zero for 100 epochs~\cite{su2018end2end}.
We additionally investigate using exponential learning rate decays with initial learning rates from \{5e-4, 5e-3\} at decay steps from \{50, 100\}.
We further use the combination of adversarial, total-variation and L1-loss of the original paper.
The first two raw measurements at phase offsets $0, \pi/2$ of the two higher frequencies 60MHz, 70MHz are used as input.

We also investigate the influence of training on image patches of resolution $128\times128$, as used by CFN and End2End originally.
During training we ensure that the cropped images inside a batch are from different scenes.

We compare the resulting MAE on the validation set after tuning and using the orignal hyperparameters (vanilla):

\begin{center}
    \begin{tabular}{c|ccc}
        & DeepToF & CFN & End2End \\
        \midrule
        vanilla & 11.97 & 4.72 & 9.14\\
        tuned   & 10.10 & 3.83 & 8.19\\
    \end{tabular}
\end{center}

The hyperparameters after tuning are:\\
DeepToF: LR 5e-4, decay 0.1 every 100 epochs, AE pre-training, full resolution. \\
CFN: LR 1e-4, decay 0.1 every 100 epochs, on patches.\\
End2End: 5e-4, decay: 0.1 every 100 epochs, full resolution.

\begin{figure}
    \centering
    \includegraphics[width=0.49\linewidth]{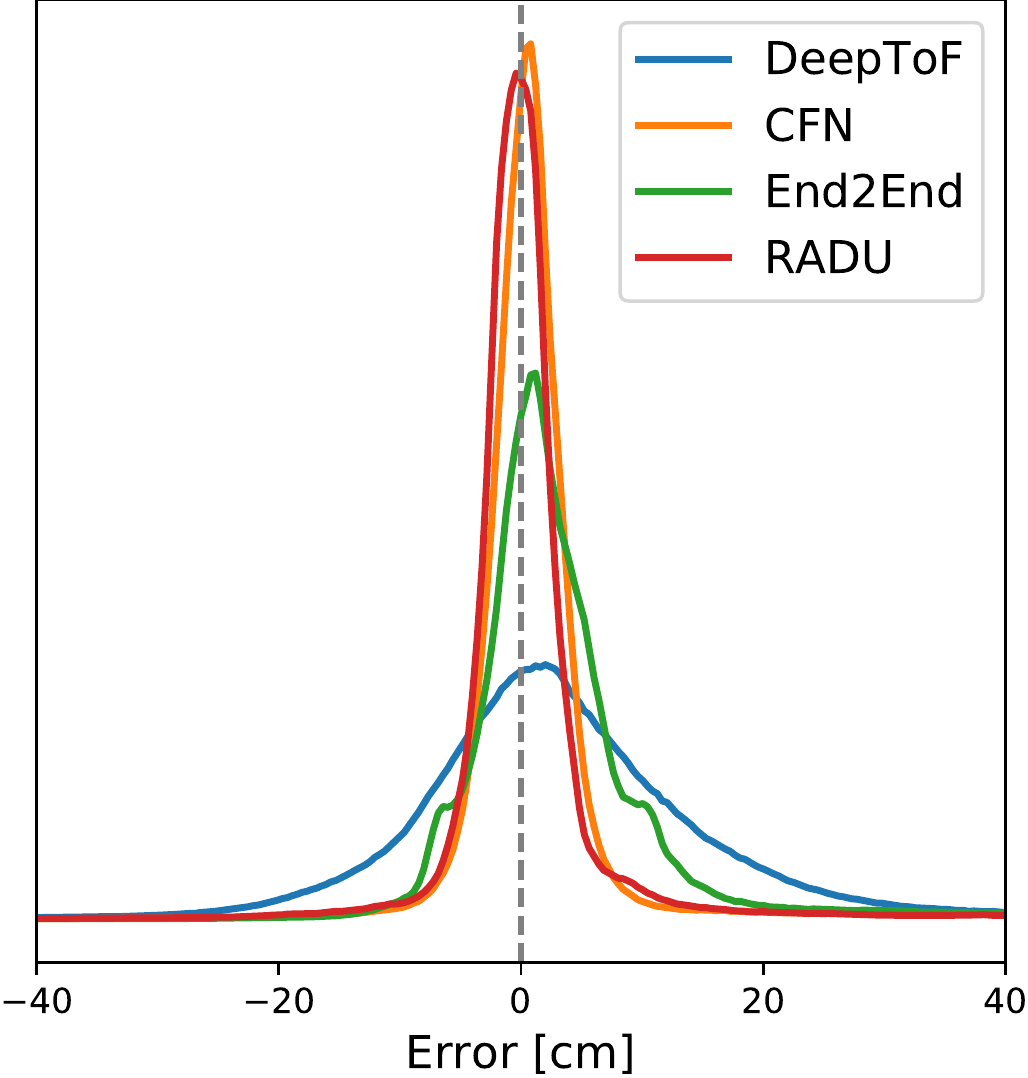}
    \includegraphics[width=0.49\linewidth]{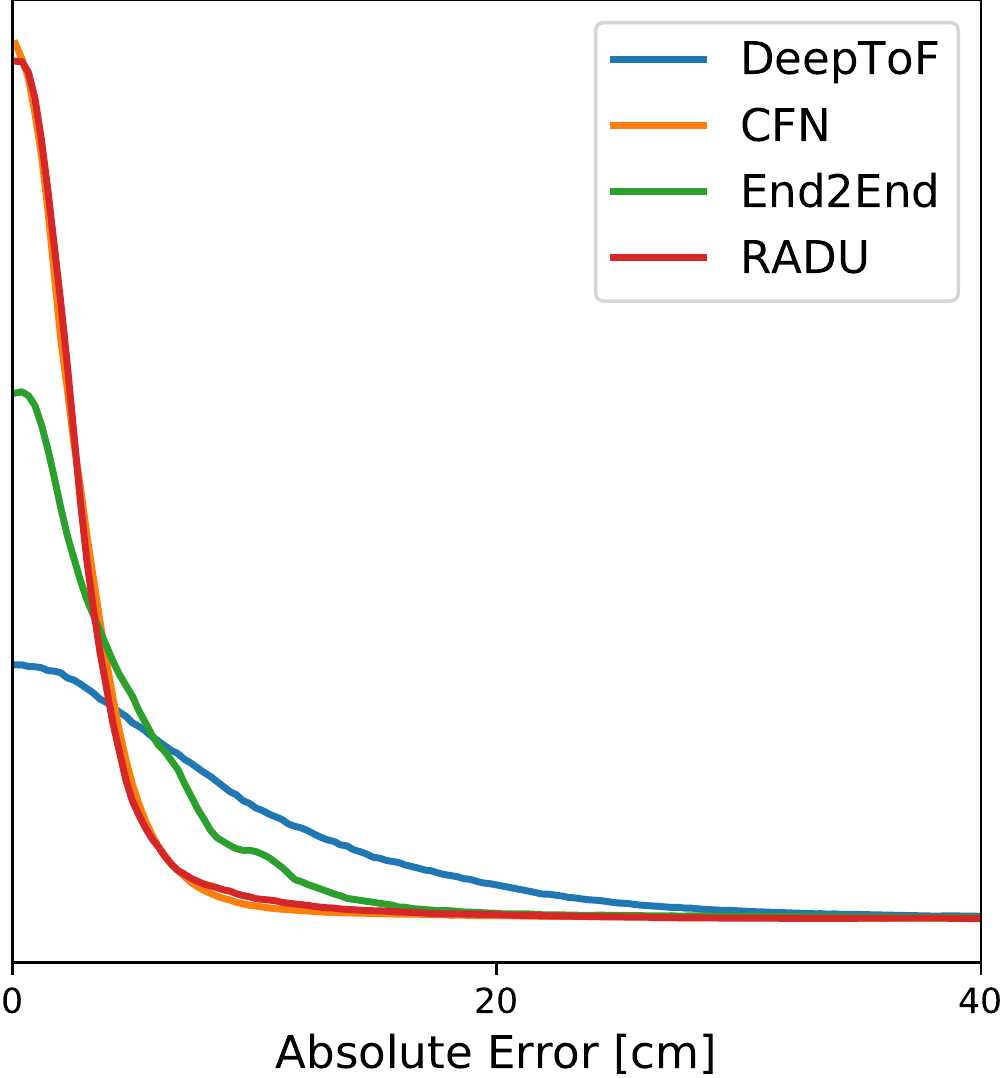}
    {\small
    \\ \ \\
    \begin{tabular}{lcc}
        \toprule
         Method &  Median &  $\sigma$\\
         \midrule
         DeepToF & 2.27 & 15.61 \\
         CFN & 0.42 & 8.80 \\
         End2End & 1.72 & 14.97 \\
         RADU & \textbf{-0.28} & \textbf{7.64} \\
         \bottomrule
    \end{tabular}
    }
    \caption{Error distributions with median and standard deviation $\sigma$ of the examined networks on our Cornell-Box dataset. An optimal distribution would be a single peak at 0. Both CFN and RADU have a similar distribution, where the median and standard deviation of CFN are slightly worse.}
    \label{fig:stats_ourdata}
\end{figure}

Error distributions and statistical values are compared in~\cref{fig:stats_ourdata}.

\subsection{Kinect2 - FLAT Dataset}

The FLAT dataset contains raw measurements simulating a Kinect2 camera for sinusoidal modulations at frequencies at 40MHz and $\thicksim$58.8MHz, and for a non-sinusoidal modulation at $\thicksim$30.3MHz.
For each modulation three measurements were performed for phase offsets with a spacing of approximately $2\pi/3$.

The resolution of the Kinect2 is $424\times 512$ which results in 3392 points in the latent point clouds of our network.
We train our network with an initial learning rate of 1e-3 and an exponential learning rate decay of 0.3 every 100 epochs, and a batch size of 2. The network converged after 600 epochs.

For DeepToF we use the low frequency ToF-depth as input.
For CFN and our network we compute the input features using $f_1=30.3\text{MHz}, f_2=40\text{MHz}, f_3=58.8\text{MHz}$.
For End2End we use the first two raw measurements of the two higher frequencies 40MHz, 58.8MHz as input.

We perform the same hyper parameter tuning as in the previous section and achieve the following MAE on the validation set:

\begin{center}
    \begin{tabular}{c|ccc}
        & DeepToF & CFN & End2End \\
        \midrule
        vanilla & 11.52 & 4.30 & 6.21\\
        tuned   & 8.66 & 3.57 & 5.90\\
    \end{tabular}
\end{center}

The hyperparameters after tuning are:\\
DeepToF: LR: 1e-3, decay: 0.1 every 50 epochs, combined training, on patches.\\
CFN: static LR: 1e-4, on patches.\\
End2End: LR: 5e-4, decay: 0.1 every 100 epochs, full resolution.

\begin{figure}
    \centering
    \includegraphics[width=0.49\linewidth]{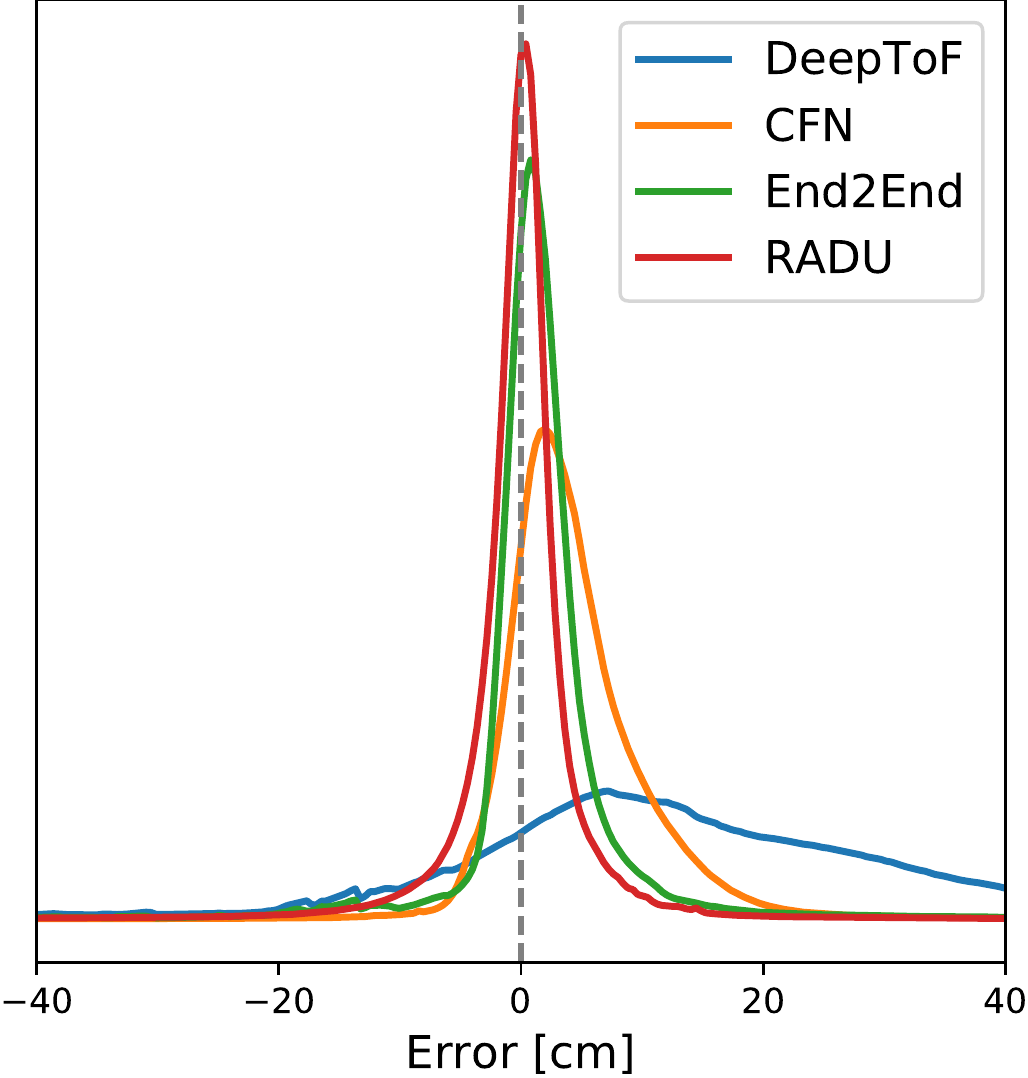}
    \includegraphics[width=0.49\linewidth]{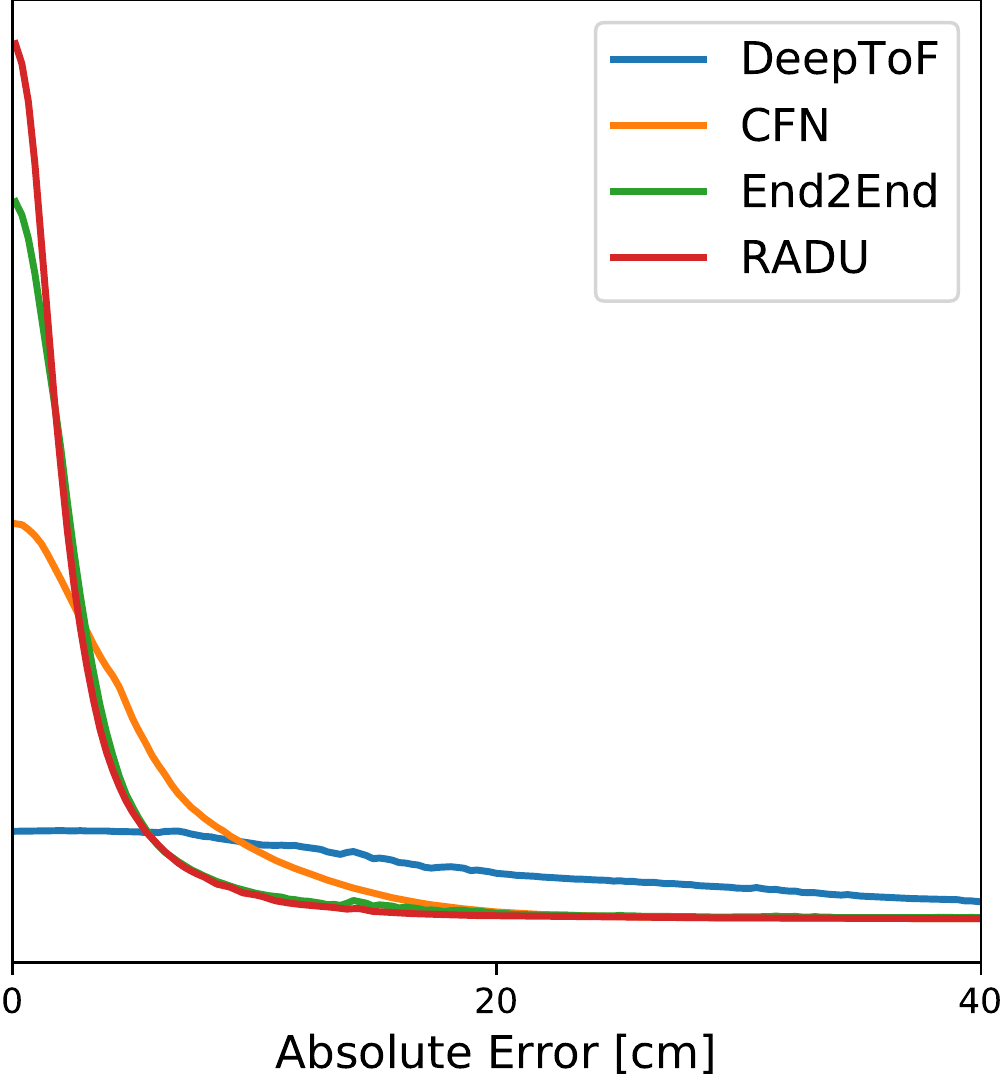}
    {\small
    \\ \ \\
    \begin{tabular}{lcc}
        \toprule
         Method &  Median &  $\sigma$\\
         \midrule
         DeepToF & 12.19& 33.08 \\
         CFN & 3.22 & 19.08 \\
         End2End & 1.07 & 25.55 \\
         RADU & \textbf{-0.08} & \textbf{7.83} \\
         \bottomrule
    \end{tabular}
    }
    \caption{Error distributions with median and standard deviation $\sigma$ of the examined networks on the FLAT dataset. An optimal distribution would be a single peak at 0.}
    \label{fig:stats_FLAT}
\end{figure}

Error distributions and statistical values are compared in~\cref{fig:stats_FLAT}.

\noindent\textbf{Code optimizations for FLAT dataset.}
The synthetic data used for training in the FLAT dataset contains a high ratio of background pixels, which we masked in the loss functions during training. 

When training on patches we ensure to have non-empty images inside the batches. 

As the background pixels are associated with depth 0, they are all projected to the same point $\vec{0}\in\mathbb{R}^3$.
This introduces a heavy memory usage in 3D convolutions as these identical points are all considered as neighbors to each other.

To reduce the memory impact we apply a filter during training, which drops all masked points in the 3D projection $\mathcal{P}_{C\rightarrow G}$, which results in a varying number of points per image.
After the 3D block of our network the points are projected back into 2D by ordering the points in a grid using a sparse data format with pixel ids inferred from the masking.

\section{Ablations}
We provide additional information about the ablation from the main paper and additional ablations on the hyperparameter $\alpha$ in our RADU convolutions, the coarse-fine-loss and on the U-DA strategy.

\subsection{Ablation 1: Latent 3D Representation}

For the 2D and 3D variants of our network architecture we use the same number of features in the 3D bottleneck block, namely $[128, 256, 128]$.
For the 2.5D convolutions, which performs 3 convolutions for foreground, neighborhood and background, we change the feature dimensions to $[129, 258, 129] = [3\cdot 43, 3\cdot 86, 3\cdot43]$, in order to make them divisible by 3.

The neighborhood radii are equal for all 3D convolutions at $[0.1m, 0.2m, 0.4m]$, for the 2D convolutions we use pixel neighborhoods of $[3, 5, 9]$, which is equal to doubling the pixel L0-distances $[1, 2, 4]$. 
For the 2.5D convolutions we use a fixed neighborhood size of $3$ pixels, as larger kernels were too demanding in memory consumption.

We use the following additional hyperparameters for the 3D convolutions:\\
KPConv: 15 kernel points.\\
PointConv: 16 hidden units in the kernel MLP.\\
MCConv: 1 kernel MLP with 16 hidden units.

We train all variants with different hyperparameters and choose the run which achieved the best validation loss.
The initial learning rate is chosen from \{1e-2, 1e-3, 1e-4\} and decayed with an exponential learning rate decay with rates \{0.1, 0.3\} every \{50, 100, 150\} epochs.
The following settings achieved the best validation MAE on S3:\\

\begin{center}
    \begin{tabular}{l|ccc}
    type & init LR & decay rate & decay steps\\
    \midrule
    2D & 1e-2 & 0.1 & 100\\
    2.5D & 1e-3 & 0.1 & 100\\
    KPConv & 1e-3 & 0.3 & 150\\
    PointConv & 1e-3 & 0.1 & 150\\
    MCConv & 1e-3 & 0.3 & 100\\
    \end{tabular}
\end{center}

\subsection{Ablation 2: Hyperparameter $\alpha$}\label{ssec:ablation_update_scale}
As discussed in the main paper the RADU layers receive direct gradients from the coarse loss, which can increase the risk of overfitting.
We investigate the influence of the regularization hyperparameter $\alpha$ of the RADU layer, and train instances of our network, again using S1 and S3, for different values of $\alpha$, including a dynamic value choice, by using the convolution radius $\alpha_l=r_l$, in our case 0.1\,m, 0.2\,m, and 0.4\,m. 
Results are reported in \cref{tab:ablation_update_scale}.
While the result indicate that a wrong choice of $\alpha$ yields the risk of overfitting on the training data, we found that $\alpha=0.1$\,m leads to good performance on the three datasets of the main paper.


\begin{table}
    \centering
    \begin{tabular}{ccc}
        \toprule
         $\alpha$ & Training (S1) & Validation (S3) \\
         $ $[m] & MAE [cm] & MAE [cm] \\
         \midrule
         0.0 & 8.38 & 2.51 \\
         0.1 & 7.87 & 2.28 \\
         0.2 & 7.51 & 2.81  \\
         1.0 & 6.93 & 3.42 \\
         $r$ & 7.48 & 2.45 \\
         \bottomrule
    \end{tabular}
    \caption{Influence of the hyperparameter $\alpha$ in the RADU convolutions. We report MAE on training (synthetic) and validation (real) data.
    The case $\alpha = r$ uses the receptive field $r$ as scale. The value $\alpha=0$ corresponds to a standard MCConv layer.}
    \label{tab:ablation_update_scale}
\end{table}

\subsection{Ablation 3: Coarse-Fine-Loss}

To investigate the influence of the coarse-fine-loss, we train an instance of our network using only the L1-Loss $\|d_{gt} - \hat{d}_{out}\|_1$, again using S1 for training and S3 for validation.

In this setting the number of epochs until the network converged during training increased by 100 epochs, while the performance on the validation set S3 decreased to a MAE of 2.63cm, +0.35 compared to the proposed coarse-fine-loss.

\subsection{Ablation 4: U-DA Strategy}

We repeat the U-DA finetuning in the real world data experiment from the main paper using an adversarial setup as proposed in other works~\cite{agresti2019unsupervised, su2018end2end}.
In this setting no pseudo-labels are created, instead a discriminator network is employed to distinguish predictions on real and synthetic data.
We implement the discriminator network and U-DA algorithm as described by Agresti \etal~\cite{agresti2019unsupervised}.

Using this setup the performance on S5 decreases to a MAE of 2.13cm (+0.5).
As with our cyclic self-training approach we repeat the training 10 times and measure a standard deviation of the MAE at 0.057cm on S5, which is notably higher (+0.036) then when fine-tuning with our proposed cyclic self-training approach using pseudo labels.

\section{Implementation}

All network implementations were done in \verb|TensorFlow 2.3.0-gpu| and \verb|Python 3.6|.
The dataset generation was done using \verb|Python 3.6| and the transient renderer of Jarabo \etal~\cite{Jarabo14transient} in version \verb|26 February 2019 - Release v1.2|.

\section{Qualitative Results}
\subsection{Cornell-Box Dataset}
We show predictions for one view point per scene in ~\cref{fig:ourdata1}, \ref{fig:ourdata2}, \ref{fig:ourdata3} and \ref{fig:ourdata4}.

We further show a larger version of Figure 8 of the main paper in~\cref{fig:RADU_step}, which shows the iterative denoising on the latent point clouds.

\subsection{FLAT Dataset}
We show predictions for a subset of the images in the dataset in ~\cref{fig:FLAT1} and \ref{fig:FLAT2}.

\begin{figure}[]
    \centering
    \includegraphics[width=\linewidth]{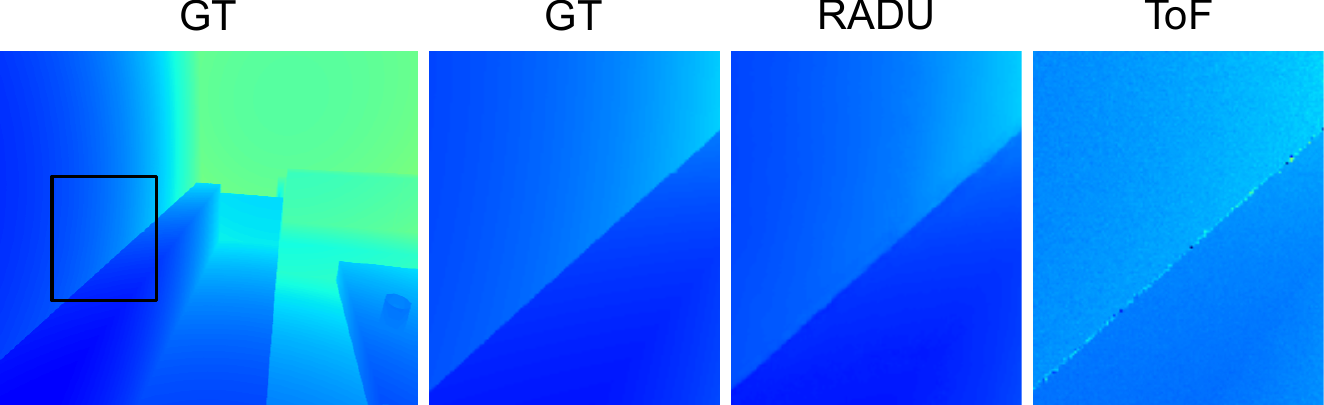}
    \caption{Example of flying pixel correction. Left shows the ground truth (GT) depth image, the right images show a zoomed in detail. 
    The ToF image exhibits flying pixels at the object boundary, which do not occur in the prediction of our RADU network.}
    \label{fig:flying_pixel}
\end{figure}

\begin{figure*}
    \centering
    \includegraphics[width=\linewidth]{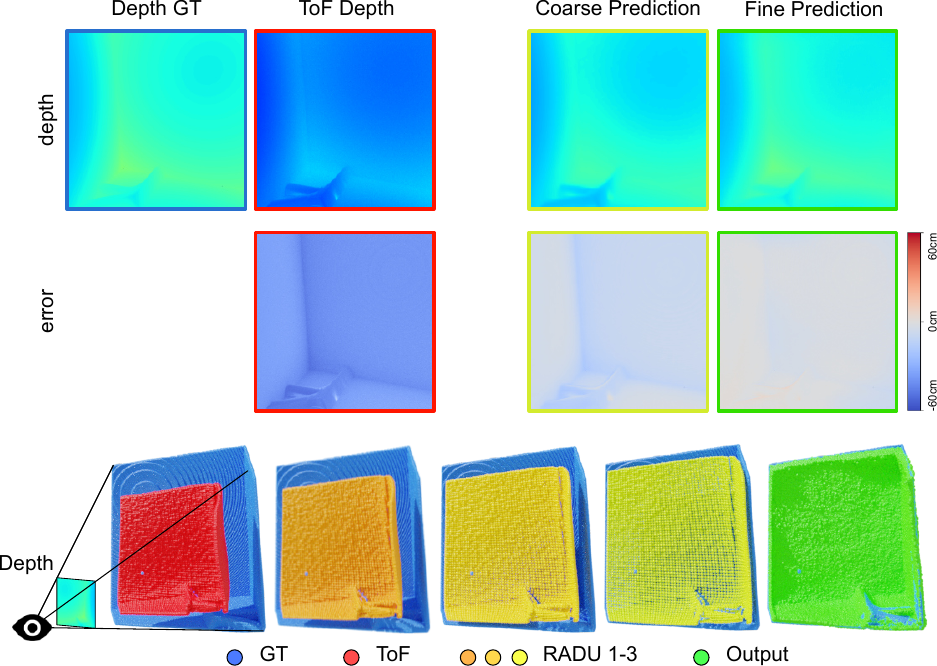}
    \caption{Larger version of Figure 8 in the main paper.
    The top row shows depth and error maps, the bottom row shows the point clouds in 3D space.
    The initial ToF depth reconstruction (red) is far from the ground truth depth (blue).
    After each RADU convolution the latent point clouds (orange to yellow) move closer to the correct depth.
    The final latent point cloud (yellow) already yields a good coarse reconstruction of the scene, which is further refined in the 2D block of the network (green).}
    \label{fig:RADU_step}
\end{figure*}

\begin{figure*}
    \centering
    \includegraphics[width=0.9\linewidth]{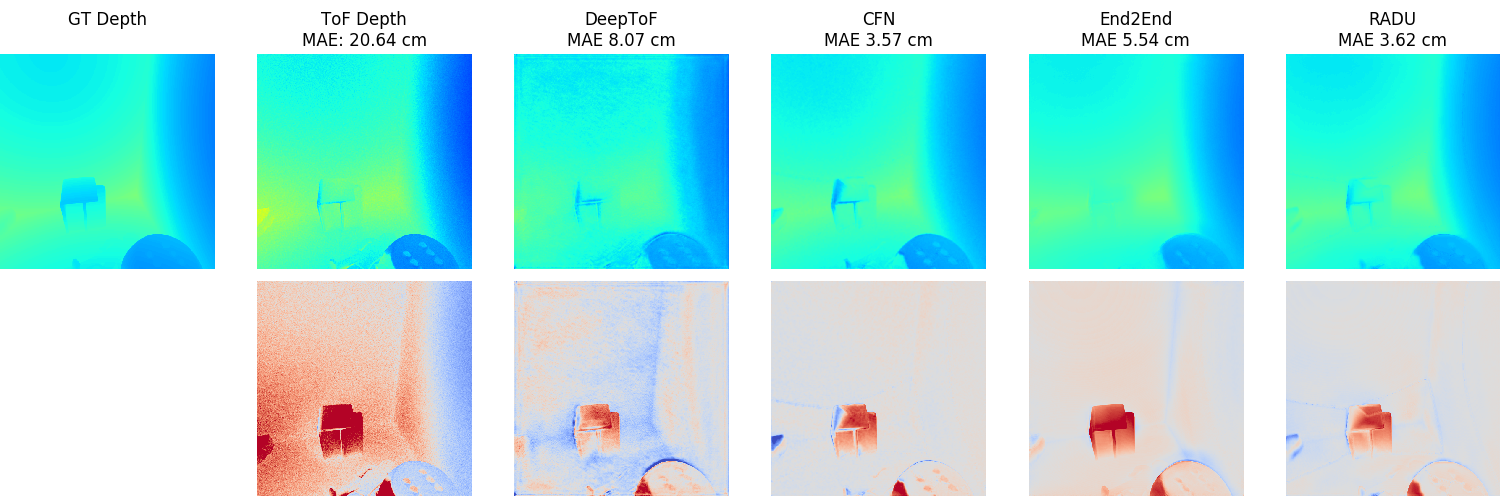}
    \includegraphics[width=0.0165\linewidth]{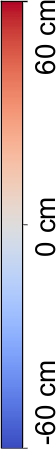}
    \caption{Results on the Cornell-Box Dataset. First row shows depths, second row shows error maps.}
    \label{fig:ourdata1}
\end{figure*}

\begin{figure*}
    \centering
    \includegraphics[width=0.9\linewidth]{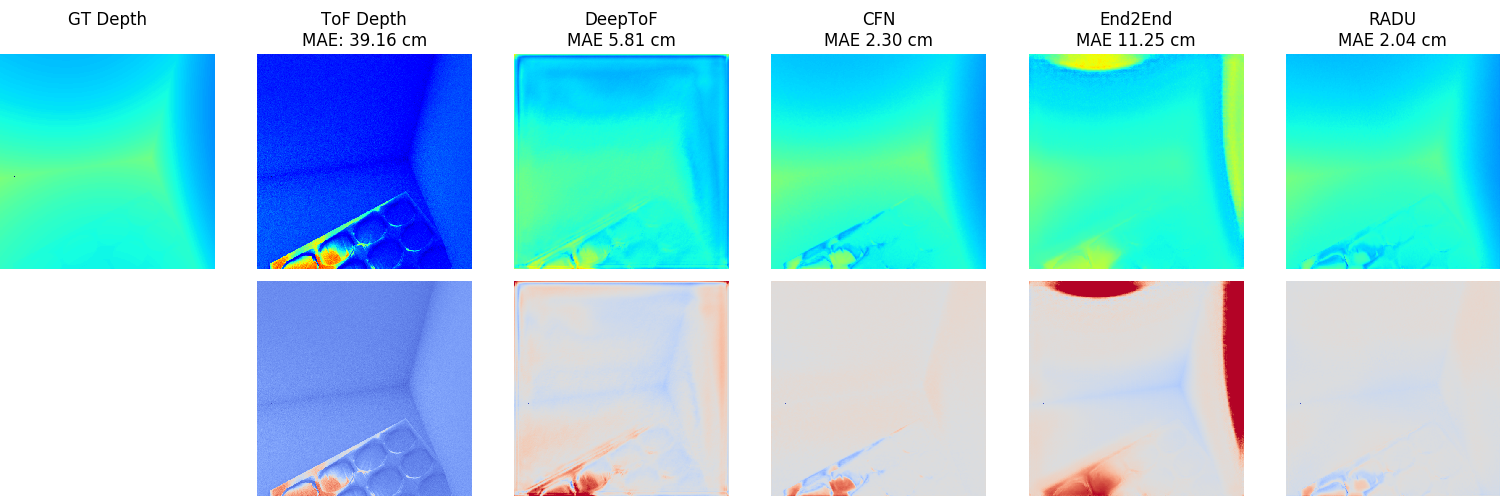}
    \includegraphics[width=0.0165\linewidth]{supp_figures/figurebar60.pdf}
    \includegraphics[width=0.9\linewidth]{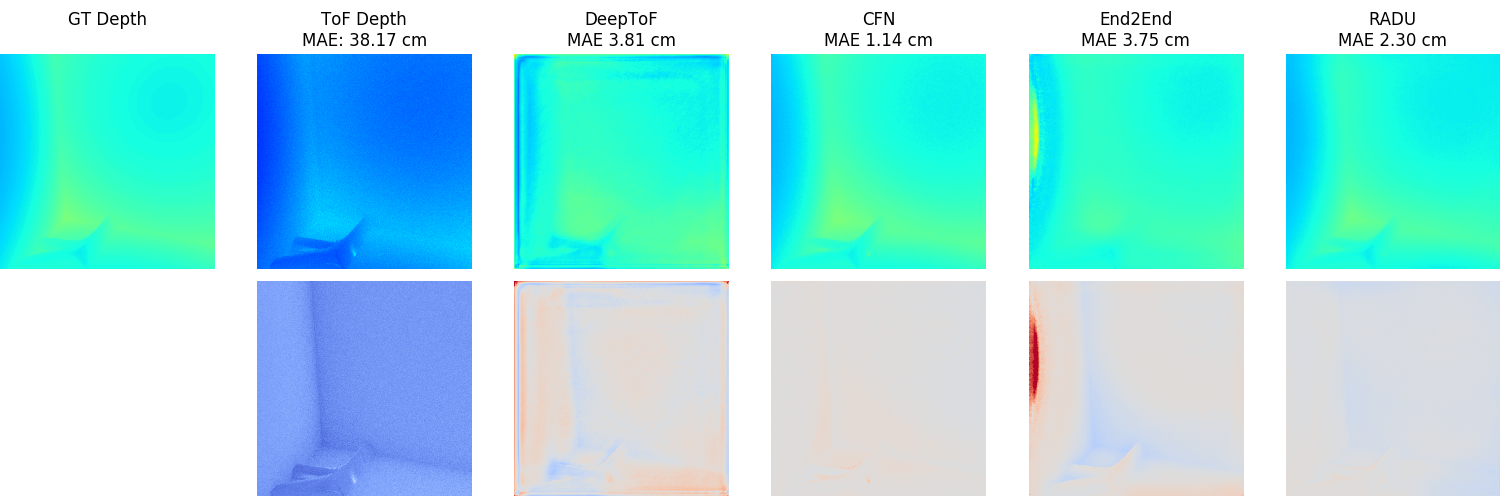}
    \includegraphics[width=0.0165\linewidth]{supp_figures/figurebar60.pdf}
    \includegraphics[width=0.9\linewidth]{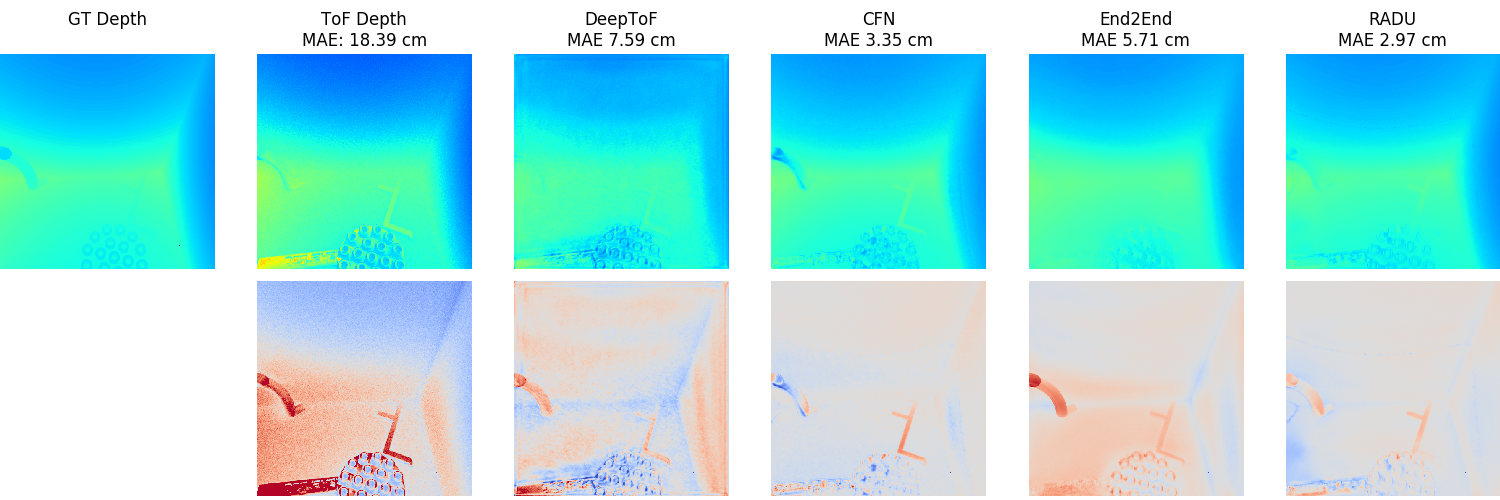}
    \includegraphics[width=0.0165\linewidth]{supp_figures/figurebar60.pdf}
    \includegraphics[width=0.9\linewidth]{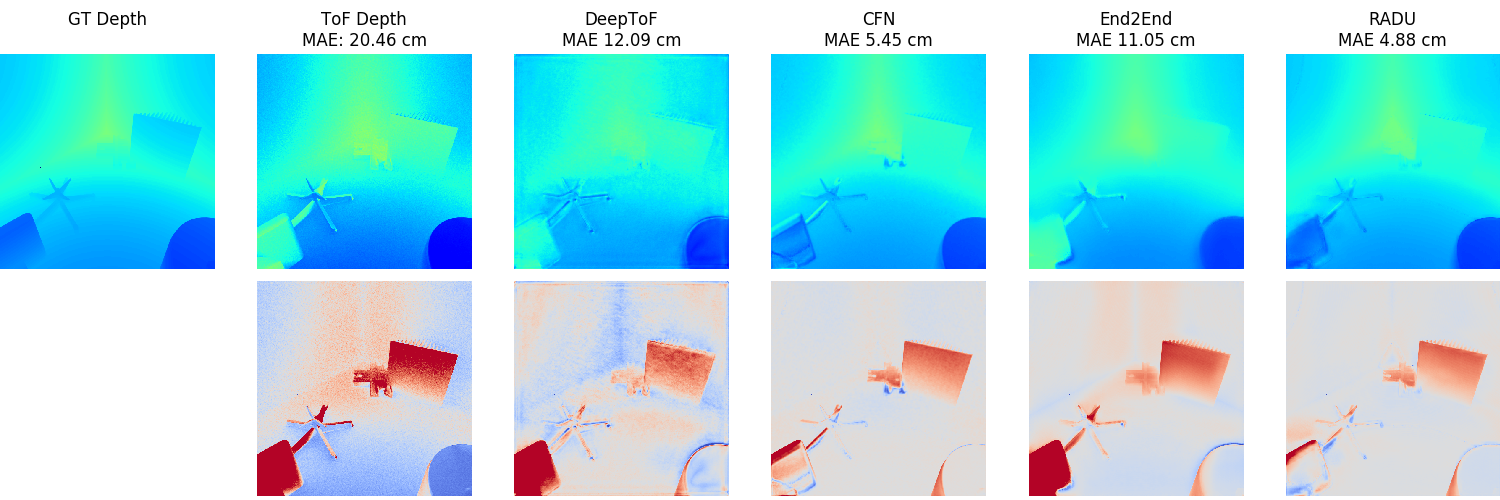}
    \includegraphics[width=0.0165\linewidth]{supp_figures/figurebar60.pdf}
    \caption{Results on the Cornell-Box Dataset. First rows show depths, second rows show error maps.}
    \label{fig:ourdata2}
\end{figure*}

\begin{figure*}
    \centering
    \includegraphics[width=0.9\linewidth]{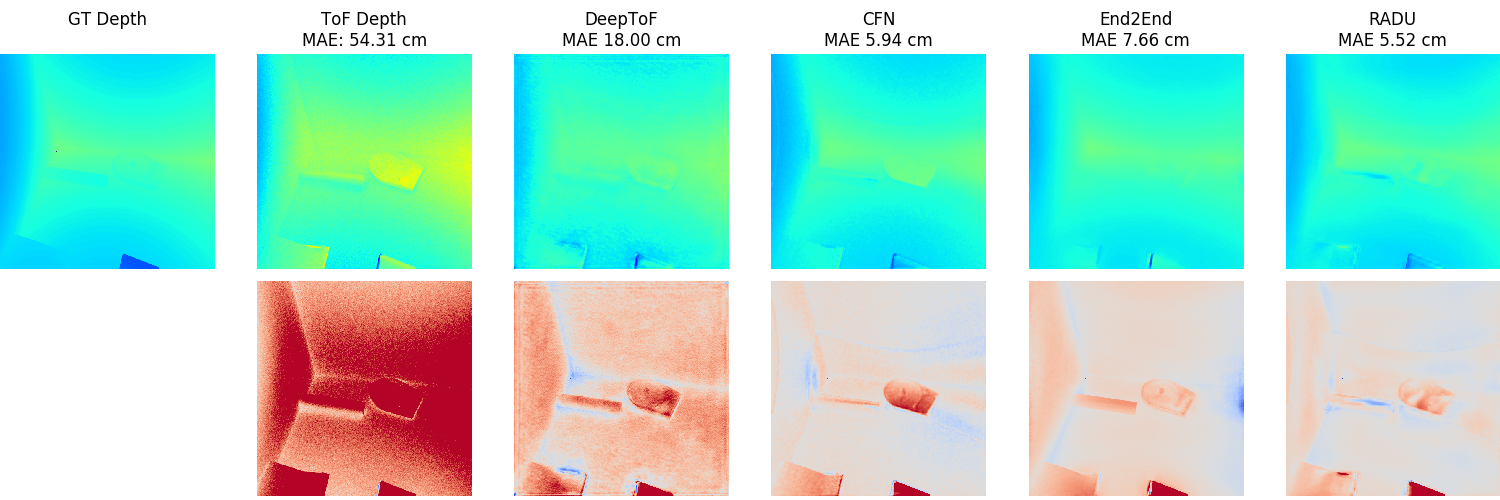}
    \includegraphics[width=0.0165\linewidth]{supp_figures/figurebar60.pdf}
    \includegraphics[width=0.9\linewidth]{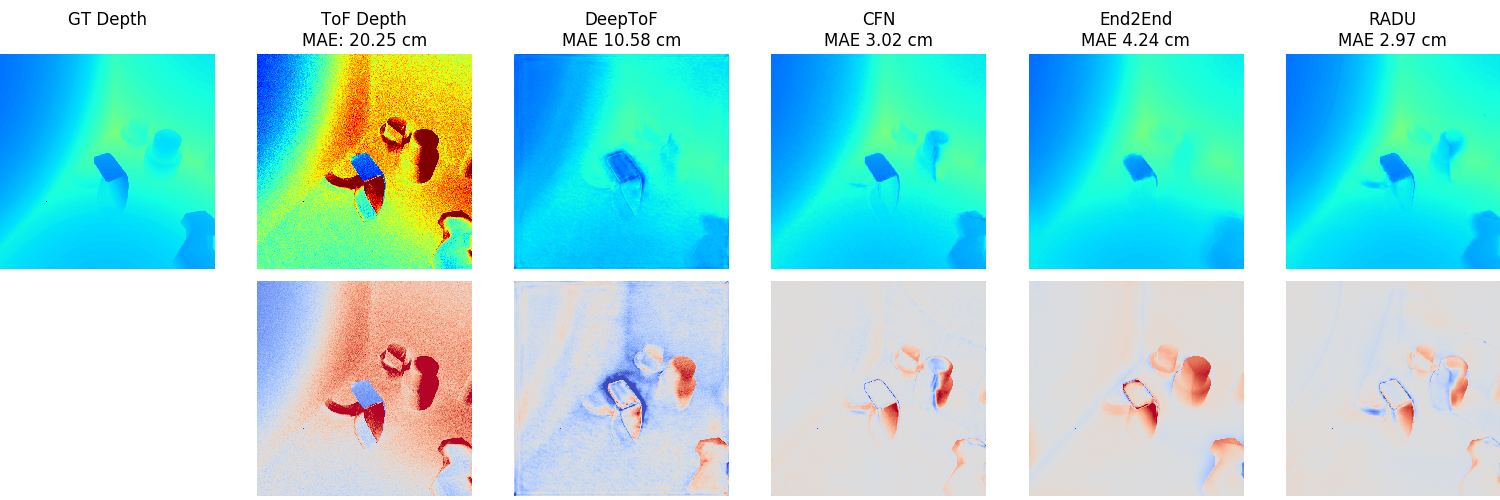}
    \includegraphics[width=0.0165\linewidth]{supp_figures/figurebar60.pdf}
    \includegraphics[width=0.9\linewidth]{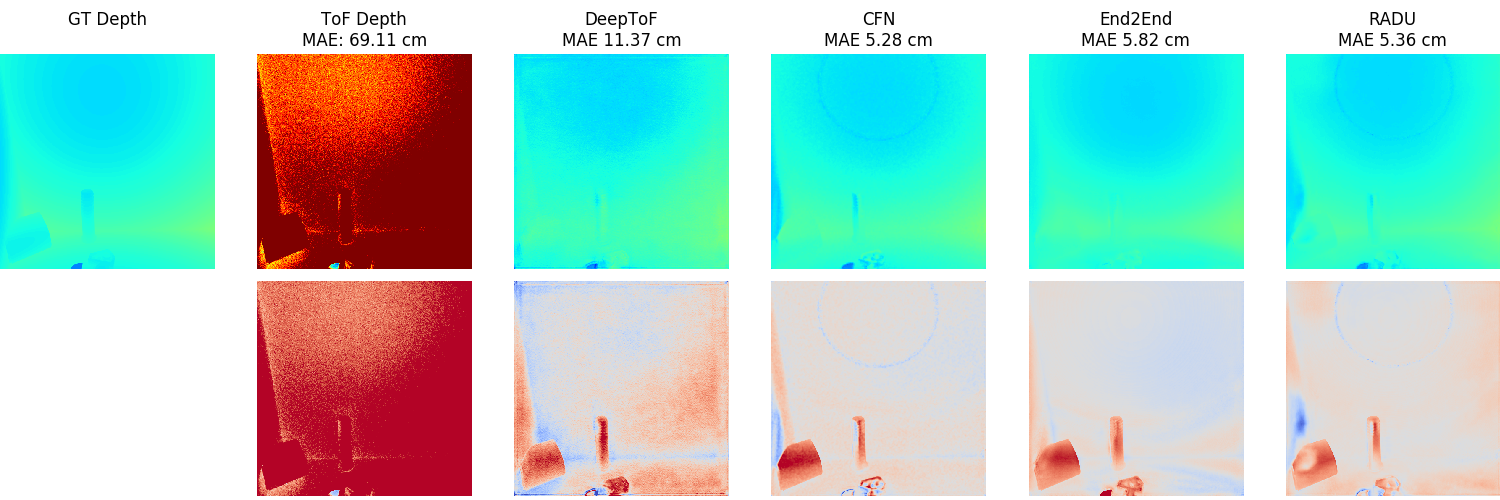}
    \includegraphics[width=0.0165\linewidth]{supp_figures/figurebar60.pdf}
    \includegraphics[width=0.9\linewidth]{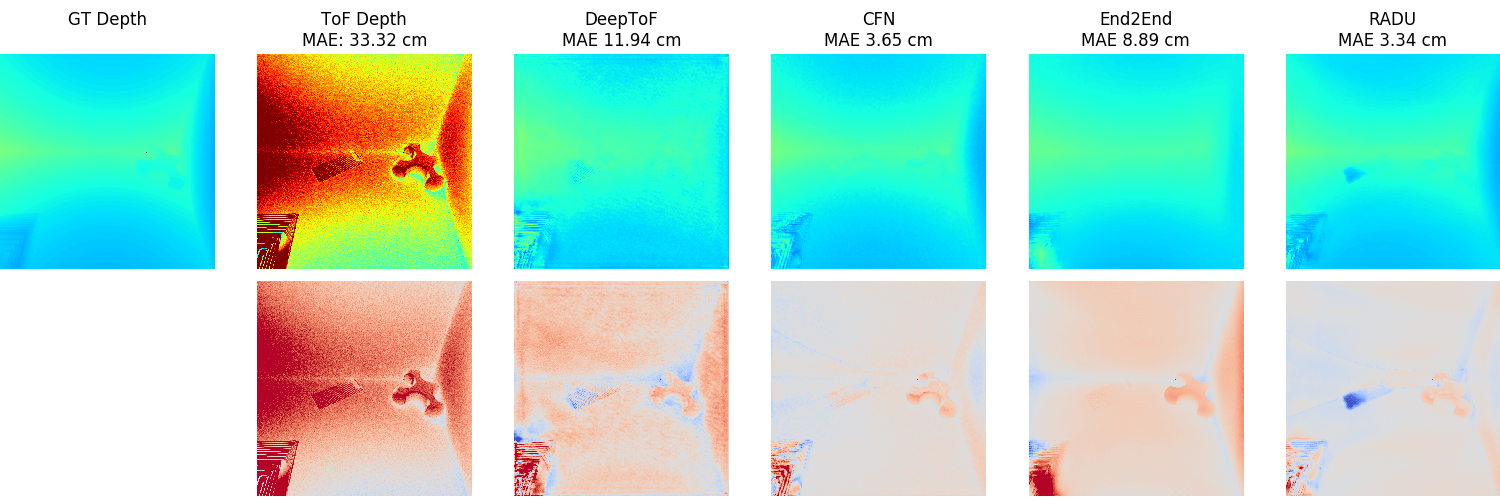}
    \includegraphics[width=0.0165\linewidth]{supp_figures/figurebar60.pdf}
    \caption{Results on the Cornell-Box Dataset. First rows show depths, second rows show error maps.}
    \label{fig:ourdata3}
\end{figure*}

\begin{figure*}
    \centering
    \includegraphics[width=0.9\linewidth]{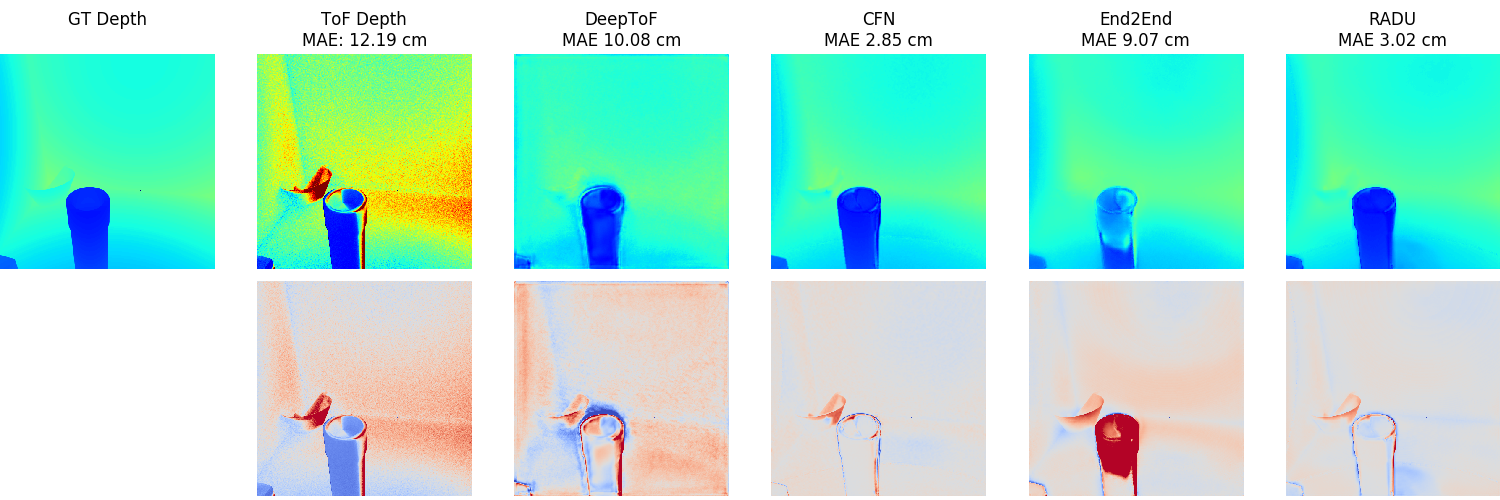}
    \includegraphics[width=0.0165\linewidth]{supp_figures/figurebar60.pdf}
    \includegraphics[width=0.9\linewidth]{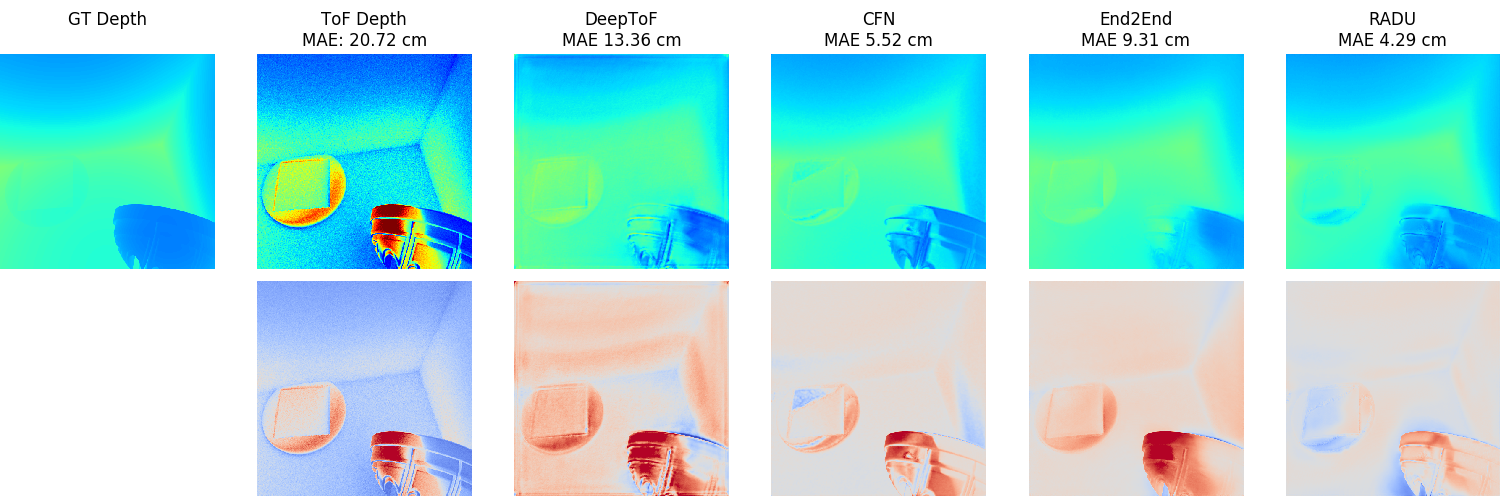}
    \includegraphics[width=0.0165\linewidth]{supp_figures/figurebar60.pdf}
    \includegraphics[width=0.9\linewidth]{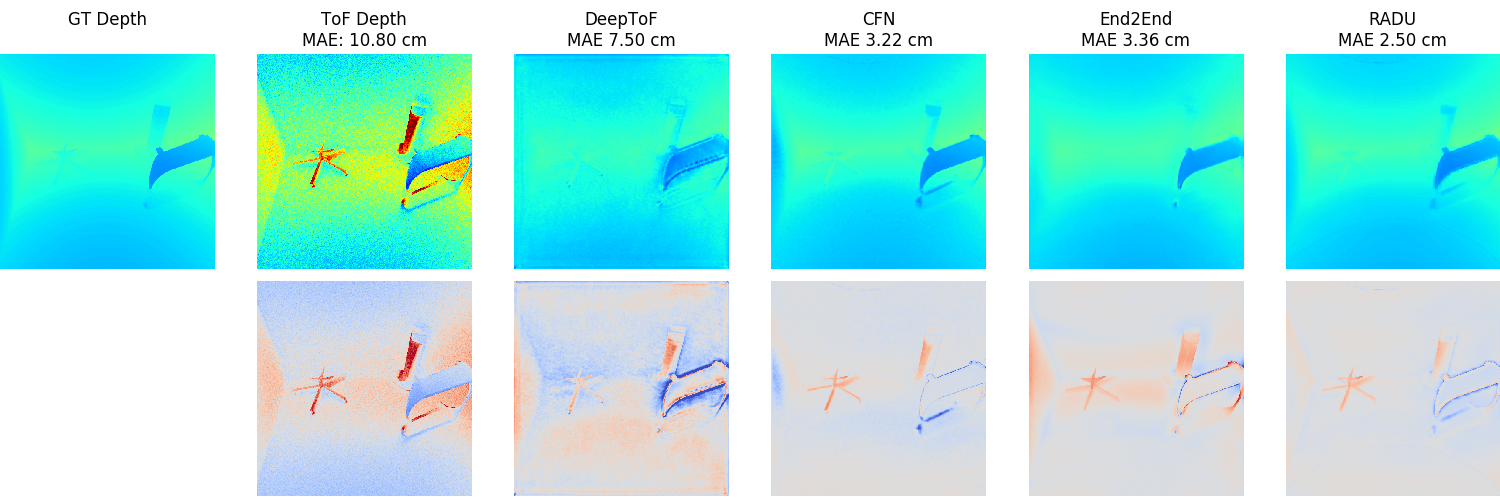}
    \includegraphics[width=0.0165\linewidth]{supp_figures/figurebar60.pdf}
    \includegraphics[width=0.9\linewidth]{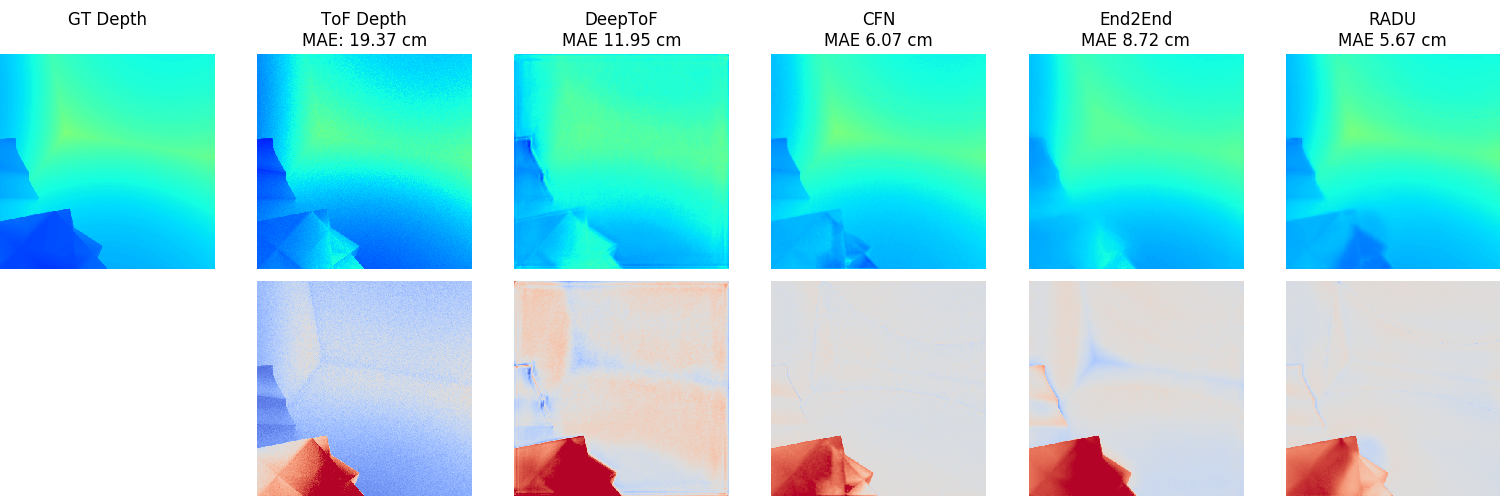}
    \includegraphics[width=0.0165\linewidth]{supp_figures/figurebar60.pdf}
    \caption{Results on the Cornell-Box Dataset. First rows show depths, second rows show error maps.}
    \label{fig:ourdata4}
\end{figure*}

\begin{figure*}
    \centering
    \includegraphics[width=0.9\linewidth]{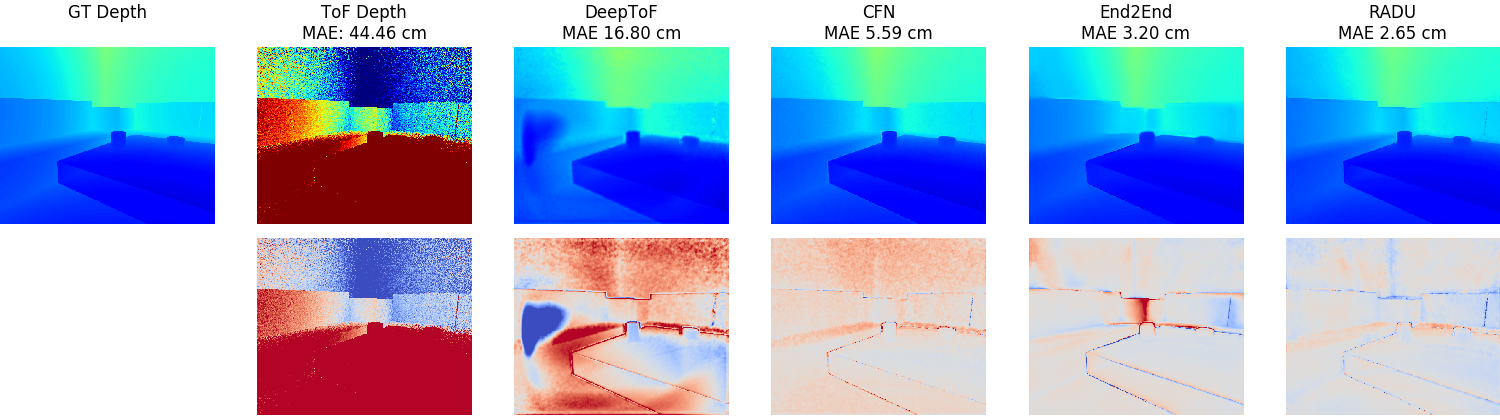}
    \includegraphics[width=0.014\linewidth]{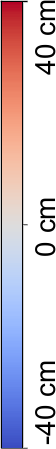}
    \includegraphics[width=0.9\linewidth]{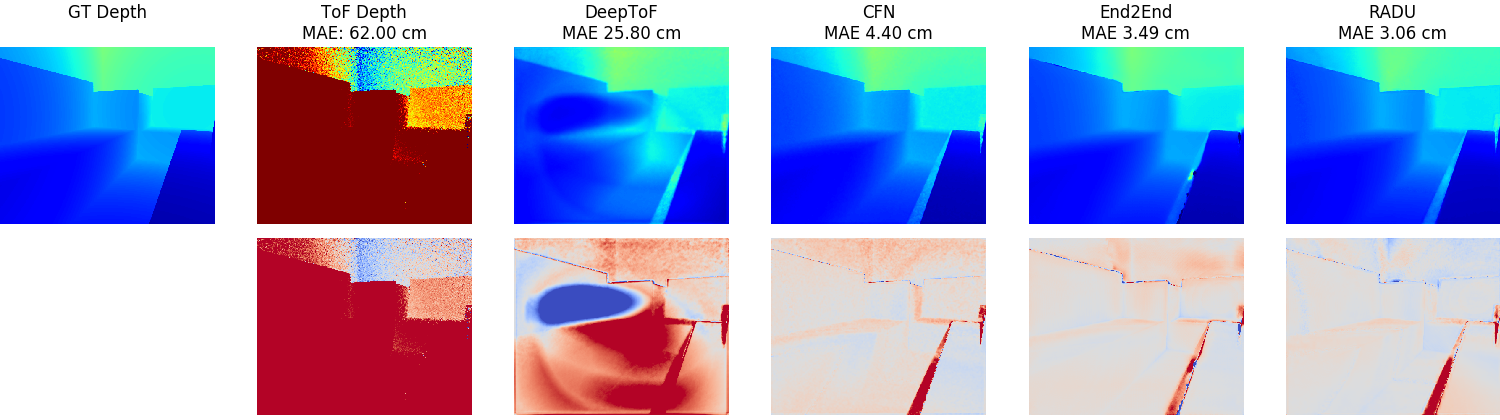}
    \includegraphics[width=0.014\linewidth]{supp_figures/figurebar40.pdf}
    \includegraphics[width=0.9\linewidth]{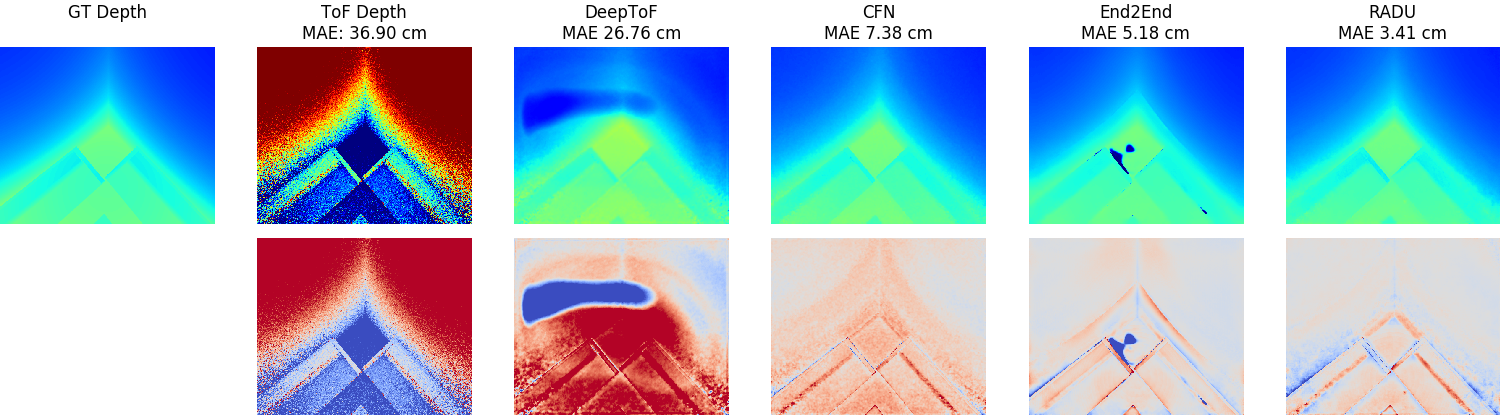}
    \includegraphics[width=0.014\linewidth]{supp_figures/figurebar40.pdf}
    \includegraphics[width=0.9\linewidth]{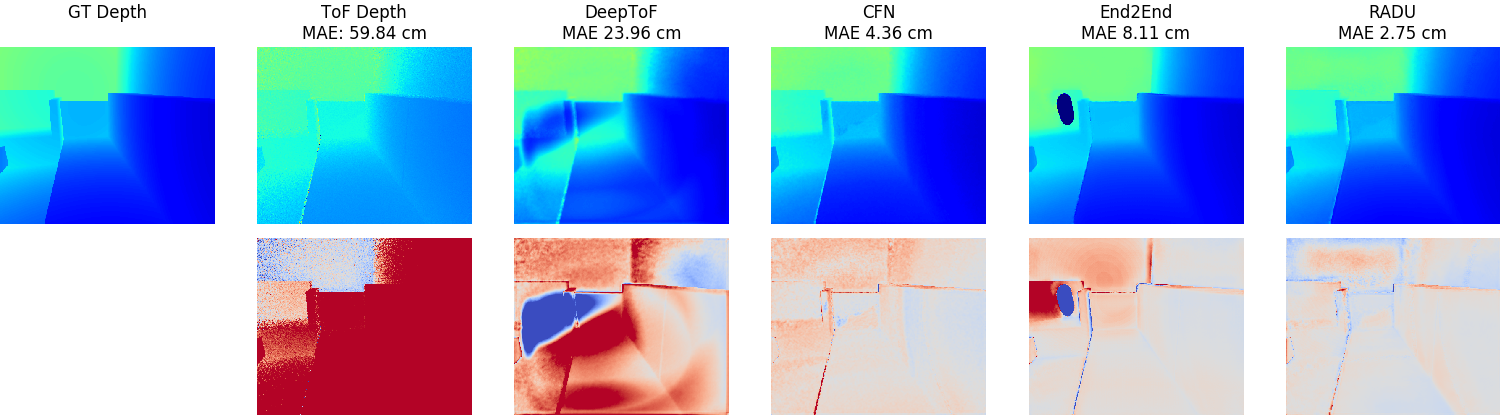}
    \includegraphics[width=0.014\linewidth]{supp_figures/figurebar40.pdf}
    \includegraphics[width=0.9\linewidth]{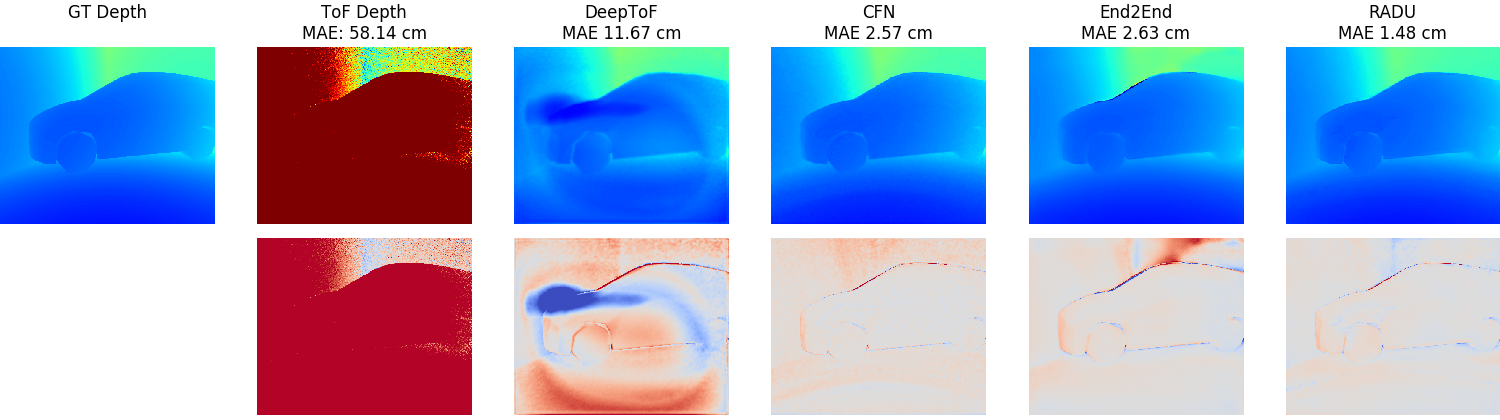}
    \includegraphics[width=0.014\linewidth]{supp_figures/figurebar40.pdf}
    \caption{Results on the FLAT Dataset. First rows show depths, second rows show error maps.}
    \label{fig:FLAT1}
\end{figure*}

\begin{figure*}
    \centering
    \includegraphics[width=0.9\linewidth]{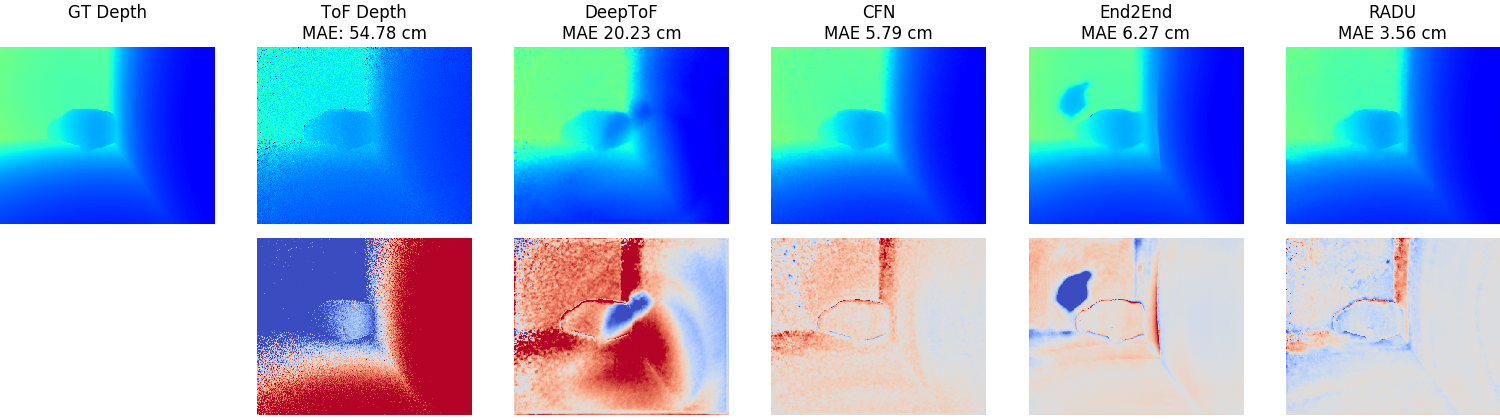}
    \includegraphics[width=0.014\linewidth]{supp_figures/figurebar40.pdf}
    \includegraphics[width=0.9\linewidth]{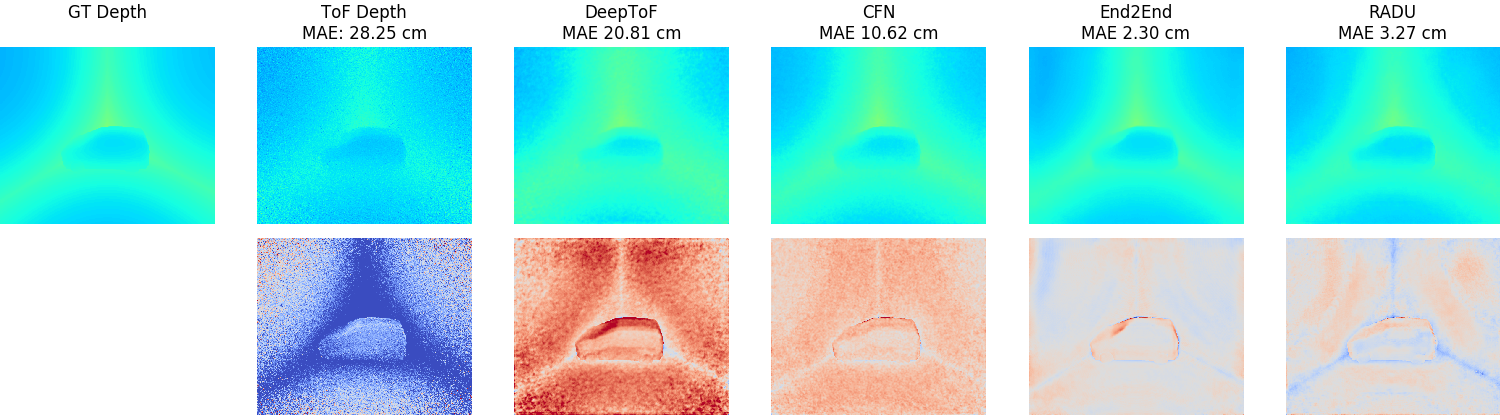}
    \includegraphics[width=0.014\linewidth]{supp_figures/figurebar40.pdf}
    \includegraphics[width=0.9\linewidth]{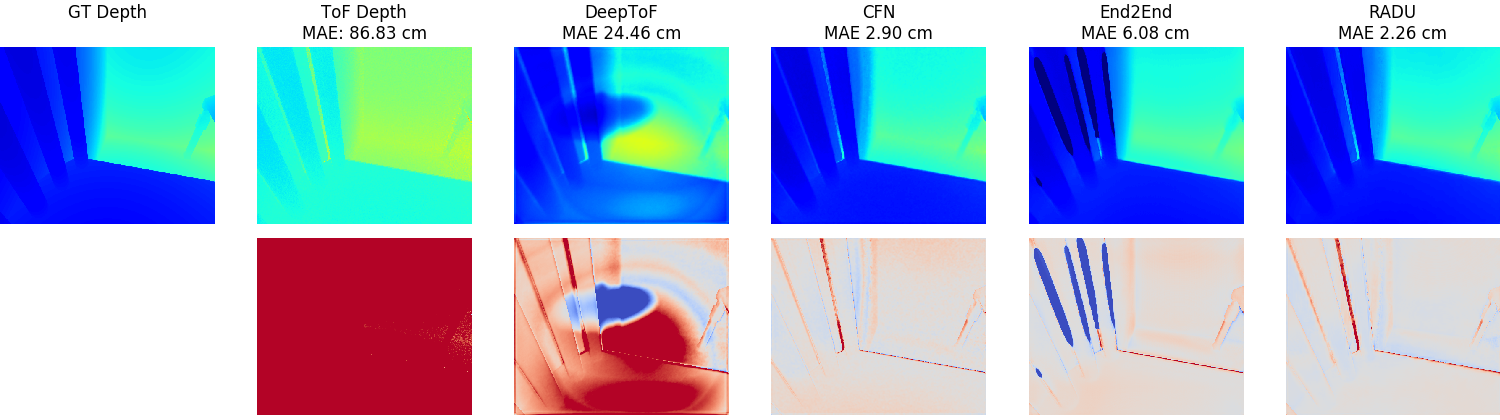}
    \includegraphics[width=0.014\linewidth]{supp_figures/figurebar40.pdf}
    \includegraphics[width=0.9\linewidth]{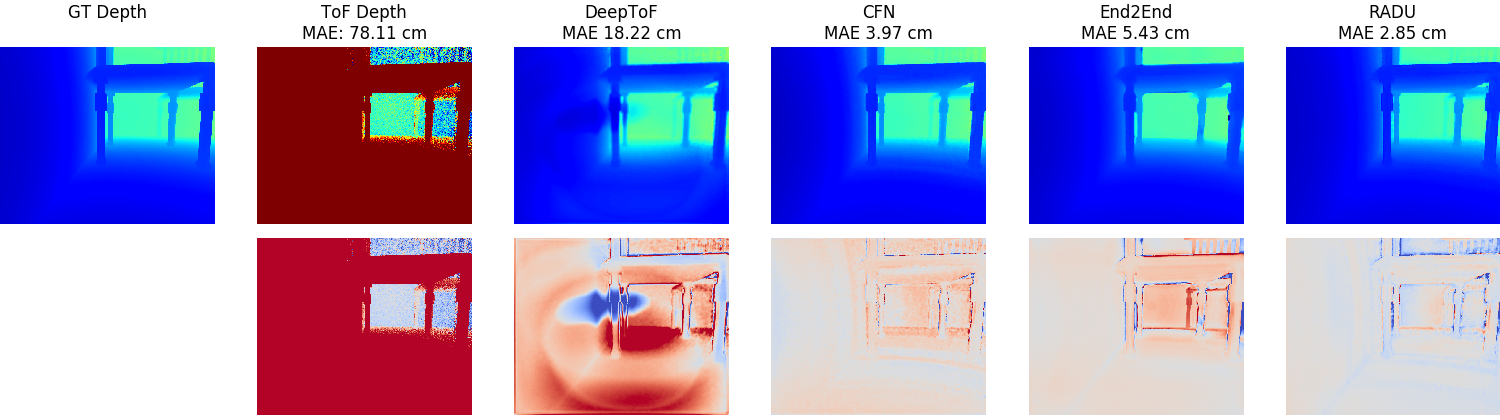}
    \includegraphics[width=0.014\linewidth]{supp_figures/figurebar40.pdf}
    \includegraphics[width=0.9\linewidth]{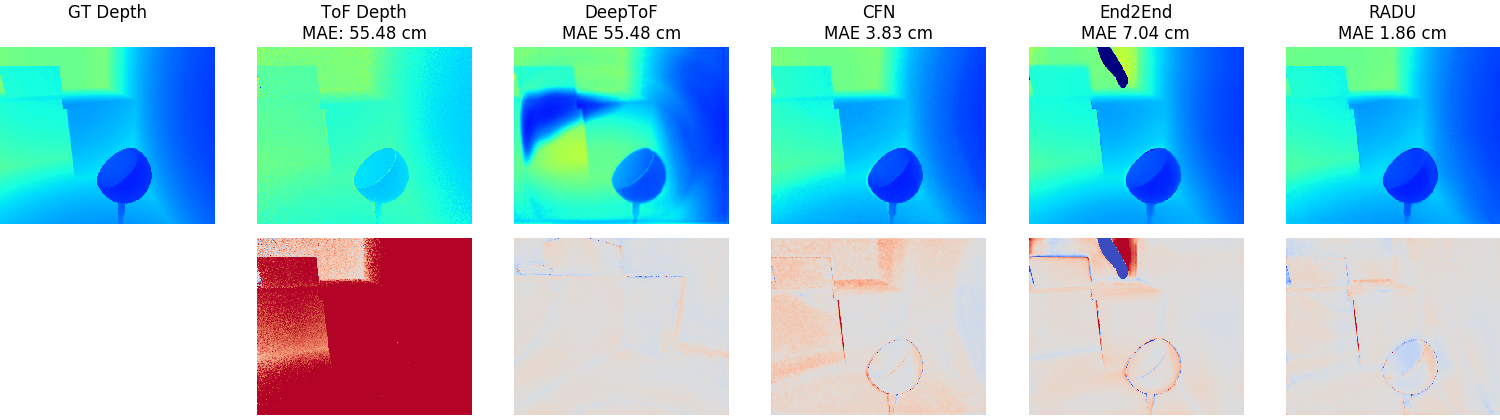}
    \includegraphics[width=0.014\linewidth]{supp_figures/figurebar40.pdf}
    \caption{Results on the FLAT Dataset. First rows show depths, second rows show error maps.}
    \label{fig:FLAT2}
\end{figure*}

\begin{figure*}
    \centering
    \includegraphics[width=0.9\linewidth]{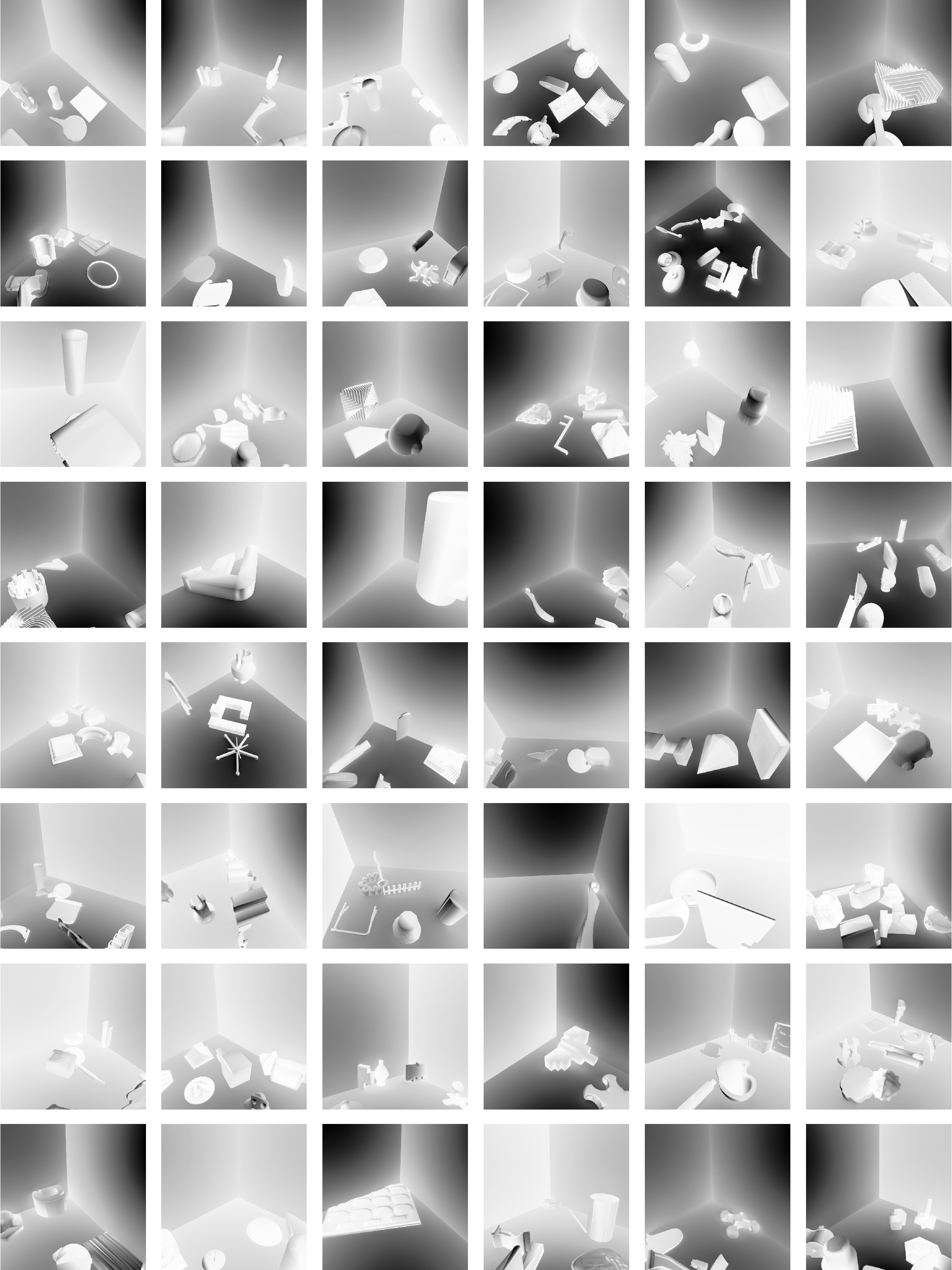}
    \caption{Example images showing one of the 50 view points with one of the three material configurations for each of the scenes in our Cornell-Box dataset. We show the intensity $I$ at a frequency of 20 MHz as given by Eq.~\eqref{eq:intensity}.}
    \label{fig:scene_table1}
\end{figure*}

\begin{figure*}
    \centering
    \includegraphics[width=0.9\linewidth]{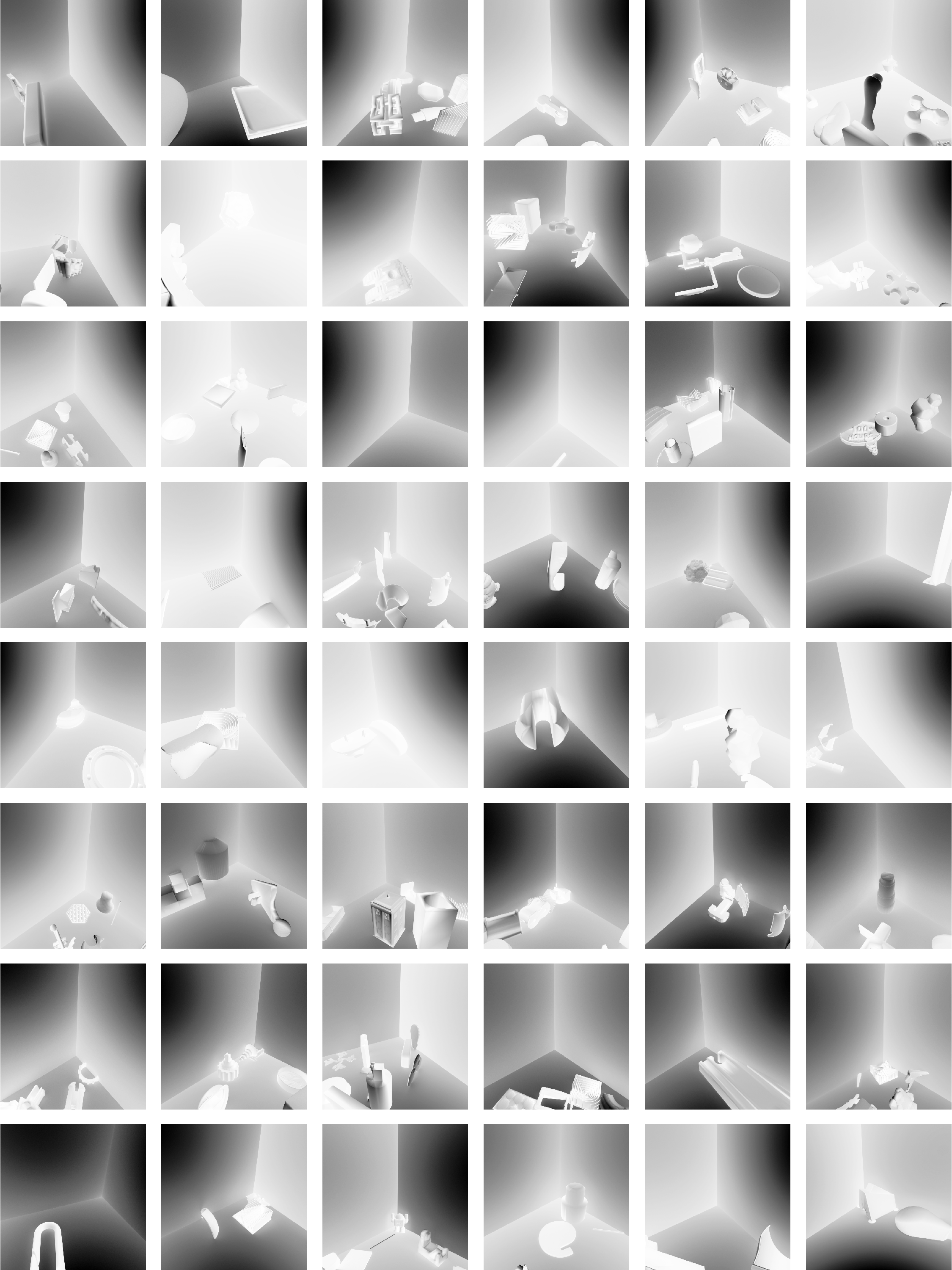}
    \caption{Example images showing one of the 50 view points with one of the three material configurations for each of the scenes in our Cornell-Box dataset. We show the intensity $I$ at a frequency of 20 MHz as given by Eq.~\eqref{eq:intensity}.}
    \label{fig:scene_table2}
\end{figure*}

\begin{figure*}
    \centering
    \includegraphics[width=0.9\linewidth]{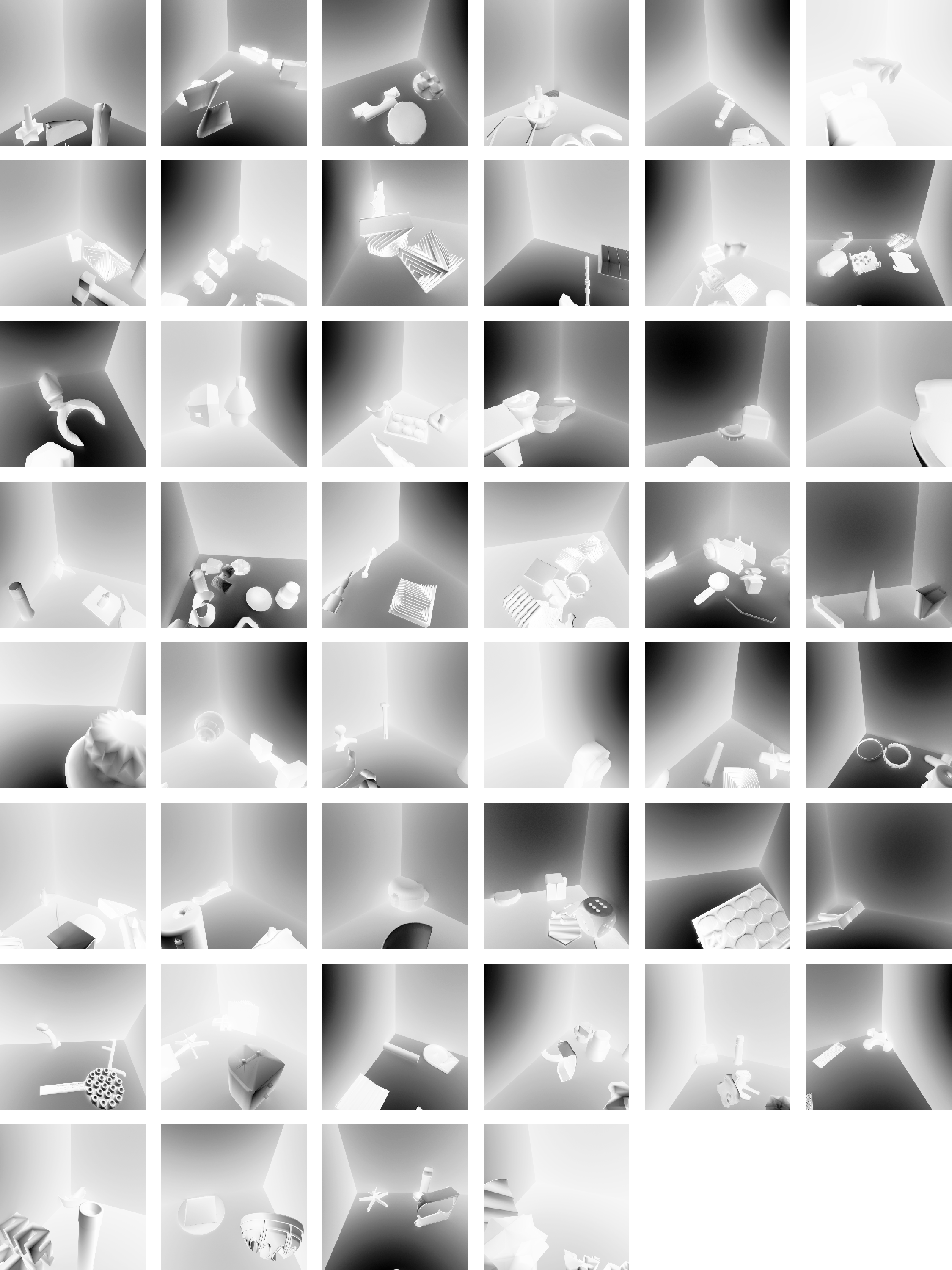}
    \caption{Example images showing one of the 50 view points with one of the three material configurations for each of the scenes in our Cornell-Box dataset. We show the intensity $I$ at a frequency of 20 MHz as given by Eq.~\eqref{eq:intensity}.}
    \label{fig:scene_table3}
\end{figure*}

Lastly, we show an example of the correction of the mixed-pixel / flying pixel effect in \ref{fig:flying_pixel}.



\end{document}